%% file: iclr2025_conference.tex
\documentclass{article} 
\usepackage{iclr2025_conference,times}

\input{math_commands.tex}

\usepackage[utf8]{inputenc} 
\usepackage[T1]{fontenc}    
\usepackage{hyperref}       
\usepackage{url}            
\usepackage{booktabs}       
\usepackage{amsfonts}       
\usepackage{nicefrac}       
\usepackage[nopatch=eqnum]{microtype}      
\usepackage{xcolor}         
\usepackage{xspace}
\usepackage{enumitem}
\usepackage{amsmath}
\usepackage{amssymb}
\usepackage{mathtools}
\usepackage{amsthm}
\usepackage{graphicx}
\usepackage{subfig}
\usepackage{float}
\usepackage{hyperref}
\usepackage{algpseudocode}
\usepackage{algorithm}
\usepackage{graphicx} 
\usepackage{makecell}
\usepackage{caption}
\usepackage{subfig}
\usepackage{wrapfig, lipsum}

\usepackage[capitalize,noabbrev]{cleveref}

\usepackage[letterpaper,margin=1in,rmargin=1in,lmargin=1in]{geometry}
\usepackage[utf8]{inputenc}

\usepackage{ebgaramond}

\crefformat{section}{\S#2\color{blue}#1#3} 
\crefformat{subsection}{\S#2\color{blue}#1#3}
\crefformat{subsubsection}{\S#2#1#3}
\setlist{topsep=0pt, leftmargin=*}

\input{defs.tex}

\title{\method: Towards Effective and Efficient External Data Augmentation}

\author{
\centerline{Zain Sarwar$^*$, Van Tran$^*$, Arjun Nitin Bhagoji$^*$, Nick Feamster, Ben Y. Zhao}\\
\centerline{University of Chicago}\\
\centerline{\texttt{\{zsarwar, tranv, abhagoji, feamster, ravenben\}@uchicago.edu}} \\
\And
\centerline{Supriyo Chakraborty} \\
\centerline{Capital One}\\
\centerline{\texttt{supriyo.chakraborty@capitalone.com}} \\
}
\iclrfinalcopy 
\begin{document}

\maketitle
\def\thefootnote{*}\footnotetext{These authors contributed equally to this work}

\begin{abstract}

   Machine learning (ML) models often require large amounts of data to perform
   well. When the data available to the model trainer is insufficient to obtain
   good performance on their desired task, they may need to acquire more data
   from external sources. Often, useful data is held by private entities who are
   unwilling to share their data due to monetary and privacy concerns. This
   makes it challenging and expensive for model trainers to acquire the data
   they need to improve their model's performance. To tackle this problem, we
   propose \method, a data-efficient method that enables model trainers to
   evaluate the relative utility of different data sources while working with a
   constrained data-sharing budget. Leveraging both functional and feature
   similarity, \method identifies small but informative data subsets from each
   data owner. This allows model trainers to identify useful data owners and
   improve model performance with minimal data exposure. Experiments across
   multiple tasks in two domains show that \method converges rapidly to the
   performance of the \fullinfo baseline, where all data is shared. Moreover,
   \method is robust to label and data noise, and can effectively recover a
   utility-based ranking of data owners. We believe \method paves the way for
   democratized training of high performance ML models.

 \end{abstract}

 \section{Introduction}\label{sec: intro}

\input{sections/intro.tex}

 \section{Problem Setup and Formulation}\label{sec: motivation} 
\input{sections/setup.tex}

 \section{\method: Identifying and sharing useful data}\label{sec: methods}
 \input{sections/methods.tex}
 
 \section{Experiments}\label{sec: evaluation}
 \input{sections/evaluation.tex}

 \section{Case Studies}\label{sec: case studies}
 \input{sections/case_studies.tex}

 \section{Related Work}\label{sec: relatedworks}
 \input{sections/relatedworks.tex}

 \section{Discussion}\label{sec: discussion}

\input{sections/discussion.tex}

\bibliography{iclr2025_conference}
\bibliographystyle{iclr2025_conference}

\newpage

\appendix
\input{sections/appendix}

\end{document}

%% file: math_commands.tex
\usepackage{amsmath,amsfonts,bm}

\def\eqref#1{equation~\ref{#1}}

\def\1{\bm{1}}
\newcommand{\train}{\mathcal{D}}

\newcommand{\test}{\mathcal{D_{\mathrm{test}}}}

\DeclareMathAlphabet{\mathsfit}{\encodingdefault}{\sfdefault}{m}{sl}
\SetMathAlphabet{\mathsfit}{bold}{\encodingdefault}{\sfdefault}{bx}{n}

\DeclareMathOperator*{\argmax}{arg\,max}
\DeclareMathOperator*{\argmin}{arg\,min}

%% file: defs.tex
\theoremstyle{plain}
\newtheorem{theorem}{Theorem}[section]

\theoremstyle{definition}
\newtheorem{definition}[theorem]{Definition}

\theoremstyle{remark}

\definecolor{ocean}{RGB}{78, 168, 245}

\definecolor{ForestGreen}{RGB}{24,155,24}

\newcommand{\problem}{external data augmentation}
\newcommand{\method}{\texttt{Mycroft}\xspace}
\newcommand{\dataowner}{data owner\xspace}
\newcommand{\modeltrainer}{model trainer\xspace}
\newcommand{\dataowners}{data owners\xspace}
\newcommand{\MT}{\texttt{MT}\xspace}
\newcommand{\MTs}{\texttt{MT}s\xspace}
\newcommand{\DO}{\texttt{DO}\xspace}
\newcommand{\DOs}{\texttt{DO}s\xspace}
\newcommand{\Dhard}{D^{\text{hard}}\xspace}

\newcommand{\Duseful}{D^{\text{useful}}_{i}\xspace}
\newcommand{\fullinfo}{\texttt{full-information}\xspace}
\newcommand{\randsamp}{\texttt{random-sampling}\xspace}
\newcommand{\dvwS}{Dogs \& Wolves - Spurious\xspace}
\newcommand{\dvwN}{Dogs \& Wolves - Natural\xspace}
\newcommand{\dataselect}{\texttt{DataSelect}\xspace}

%% file: sections/intro.tex
Machine learning models' performance relies heavily on their training datasets,
but the data available for training is often insufficient, outdated, or
unrepresentative of the task \citep{9234592, elsahar2019annotate,
10.1145/3609422}. As models expand into new domains and existing ones exhaust
public data, data scarcity becomes a key challenge. According to economic and
manufacturing experts, one of the primary reasons that AI technology is not
being widely adopted in manufacturing is the lack of relevant public data
available for production tasks \citep{alam2024automation}. While large
corporations and governments can afford large-scale data collection or use
methods like crowdsourcing~\ \citep{sigurdsson2016hollywood} and federated
learning \citep{kairouz2021advances} to increase their access to diverse
datasets, entities without such resources, such as individuals and small
businesses, find it extremely challenging to collect their own data.

Such entities may need to acquire data from private entities (referred to as
\emph{\dataowners}), who are often unwilling to share it publicly, especially
when it involves proprietary or sensitive data, such as educational or health
records \citep{spector2019genetic}. Acquiring data from these private
\dataowners can involve high overhead, such as the creation of complex data
sharing agreements, monetary compensation, and compliance with regulations such
as \citep{GDPR_2021}. Therefore,
data sharing protocols which help model trainers assess which data owners are
most likely to provide useful data before acquiring data at scale are needed.






\noindent \textbf{Challenges of external data augmentation:} The \modeltrainer
 needs to efficiently select one or more \dataowners whose datasets match their
 requirements. In the case with a profusion of \dataowners, while a large number
 can be filtered out just on the basis of metadata (\textit{i.e.} domain,
 collection methodology, licensing requirements etc.), it is unclear which of
 the remaining \dataowners~the model trainer should acquire data from. Even in
 the case of a single \dataowner~with a large pool of data, both the
 \modeltrainer~and \dataowners~will benefit from a method to assess the
 dataset's potential effectiveness. In summary, the key question we aim to
 answer in this paper is:

\textit{How can a model trainer assess the usefulness of data owners without acquiring their entire dataset?}

\noindent \textbf{\method for external data augmentation:} We present \method, a
framework that helps the \modeltrainer and \dataowners identify relevant data
(from the \dataowners)  
for improving the \modeltrainer's performance. The framework follows three main
steps: (1) the \modeltrainer sends information about the task on which their
model is under-performing, typically through data samples, to \dataowners; (2)
the \dataowners use an efficient algorithm to identify a small, relevant subset
of their data to demonstrate their dataset's usefulness and send it to the
\modeltrainer; and (3) the \modeltrainer~evaluates this subset to decide whether
to acquire additional data from the \dataowners.

The \emph{key technical challenge} we solve in this paper lies in Step (2),
where each \dataowner~must find the most ``relevant'' training data with respect
to the model trainer's transmitted data samples, which may not follow the same
distribution as their training data. Our proposed algorithms select data based
on either functional similarity, which compares sample-wise gradients from
task-specific models, or feature similarity, which measures the distance between
samples in a relevant feature space, or a combination of both. In summary, our
contributions are as follows:





\noindent \textbf{1. Modelling the problem of external data augmentation
(\cref{sec: motivation}:)} We formally model the practical problem of acquiring
data from private data owners by stating key assumptions and adding constraints
inspired by the likely real-world operating conditions for any external data
augmentation algorithm.

\noindent \textbf{2. \method: a protocol for identifying and sharing useful data
(\cref{sec: methods}:)} We outline a data sharing protocol, which crucially
depends on methods for the \dataowner to identify relevant data. We propose two
data selection approaches, based on functional (loss gradient) similarity and
feature similarity, and highlight their applicability. We adapt existing tools
to make them suitable for our goal and in cases where these techniques were
ineffective (such as for tabular data), we develop a new metric for data
selection. Finally, we unify both methods via a joint optimization objective to
leverage the notions of similarity arising from each approach.

\noindent \textbf{3. Extensive methodological assessment (\cref{sec:
evaluation}:)} We compare \method against two baselines: (i) \fullinfo, where
each \dataowner shares all their data with the \modeltrainer, serving as the
optimal baseline, and (ii) random uniform sampling. Across multiple
classification tasks from 2 domains (computer vision and tabular data) and a
range of data budgets, our experiments show that \method~significantly
outperforms random sampling and quickly approaches the performance of \fullinfo
under a much smaller budget. In the vision domain, \method outperforms \randsamp
by an average of 21\% for five data budgets on four datasets. In the tabular
domain, \method with just 5 samples outperforms \randsamp with 100 samples and
matches \fullinfo performance for 65\% of sharing scenarios.


\noindent \textbf{4. Case studies of using \method in practical settings
(\cref{sec: case studies}):} We conduct four additional case studies to evaluate
\method's effectiveness across different scenarios that may be encountered in
practice. We show that \method is robust to instance and label noise in the data
provided by the \dataowners. When this data lacks labels entirely, feature similarity
is still effective at finding relevant data samples. When multiple data
providers are involved, \method successfully reconstructs a utility-based
preference order that closely aligns with the \fullinfo setting. These findings
confirm that \method offers a highly data-efficient solution for \dataowners to
reliably demonstrate the value of their data to a \modeltrainer, even in noisy
or complex environments.

We hope \method and its associated open-source
code~\footnote{Code is available at :
https://github.com/zainsarwar865/Mycroft-Data-Sharing} paves the way for more
efficient and private data-sharing frameworks, in turn helping democratize the
training of performant ML models.

%% file: sections/setup.tex
In this section, we first present the challenges associated with external data 
augmentation in practical settings and the desired properties for any proposed
method. We then formally define the specific problem we solve and present the
required notation.


\subsection{Challenges of external data augmentation}\label{subsec: challenges}

The most direct approach to external data
augmentation is to acquire all data from all available data providers. However,
this approach presents the following challenges:
\begin{enumerate}

    \item \textbf{Data owners are unwilling to share all their data:}  Due to privacy and 
    proprietary concerns, data owners are often unwilling to share all their data. 
    They also typically expect compensation which necessitates the use of
    data budgets to establish the utility of the dataset before sharing.

    \item \textbf{Sharing poorly curated data can negatively impact performance:} A large portion 
    of the data owned by \dataowners may be poorly curated or irrelevant for model training. Including 
    such data in the training process can degrade model performance, 
    so it is crucial for model trainers to select only high-quality and relevant subsets from the data owners' contributions.
    \item \textbf{Data budgets can ease computational concerns:} Sharing large 
    volumes of data from multiple owners makes it difficult and costly for model 
    trainers to update their models frequently.

\end{enumerate}


\subsection{Problem Formulation and Notation}\label{subsec: notation}

We consider a setting where there is a model trainer (\MT) and $m$ data owners
(\DOs). \MT~has trained a model $M_{\text{MT}}$ on its own dataset
$D^{\text{MT}}$ and is aiming to improve their performance (or lower their loss)
on test data $D^{\text{test}}$ via \problem. The performance of \MT's model on
$D^{\text{test}}$ is measured with respect to some task. In this paper, we
assume that the task is supervised learning, so the overall performance is
measured as $L(D^{\text{test}},M) = \sum_{z_i \in D^{\text{test}}}
\ell(y_i,M(x_i))$, where $z_i=(x_i,y_i)$ is a labeled sample and
$\ell(\cdot,\cdot)$ is some appropriate loss function such as
cross-entropy loss.

We then posit that there exists a subset $\Dhard$ of $D^{\text{test}}$ on which
\MT~is aiming to improve their performance, leading them to use \problem. We
explain how $\Dhard$ is constructed in specific settings in~\ref{subsec:
exp_setup}. Each \DO~has a dataset $D_i$ which could aid \MT~in improving their
performance on $\Dhard$. However, due to the challenges highlighted in
\cref{subsec: challenges}, the \DOs~do not share all of their data with \MT.
Rather, they share a small subset $D^{\text{useful}}$ of up to size $k$, which
we call the \emph{budget}. If $D^{\text{useful}}$ is able to improve the
performance of $\MT$'s model, then \MT~and \DO~could potentially enter into a
data-sharing agreement for additional data acquisition. This paper focuses on
how each \DO~can identify $D^{\text{useful}}$ and subsequently, how \MT~can
utilize this data to improve performance and if $m>1$, rank the \DOs. The task
for each \DO~is then:

\begin{definition}[Task for each \DO]
    Find $ D^{\text{useful}}_i \subseteq D_i $ such that $|D^{\text{useful}}_i|
    \leq k$ and $ L(\Dhard,M'_{\text{MT}}) \leq L(\Dhard,M_{\text{MT}}) $, where $M_{\text{MT}}
    = \train(D^{\text{MT}})$ and $M_{\text{MT}}' = \train(D^{\text{MT}} \cup
    \Duseful)$.
\end{definition}

We summarize our key assumptions below about the data held by various entities:
\begin{enumerate}[start=1,label={\bfseries A.\arabic*}]
    \item There exists a subset $\Dhard \subseteq D^{\text{test}}$ with accurate
    ground-truth labels on which $M_{\text{MT}}$ performs poorly, which
    \MT~shares with the \DOs.
    \label{a1}
    \item Each participating \DO~has samples from at least one of the classes
    contained within $\Dhard$.\label{a2}
\end{enumerate}

We assume~\ref{a1} because if \MT does not share any knowledge of the difficult
subset, \DOs~cannot share meaningful data. Further, if $\Dhard$ is incorrectly
labeled, basic challenges regarding performance evaluation arise.
~\ref{a2} just rules out \DOs~with no relevant data to the task under
consideration.

Additionally, our algorithm accounts for the following constraints
likely to be encountered in practice:
\begin{enumerate}[start=1,label={\bfseries C.\arabic*}]
    \item \MT~does not share $M_{\text{MT}}$ with the \DOs.~\label{c1}
    \item No \DO~ shares their full training data with \MT, \emph{i.e.}
    $ \forall \, i, k<|D_i|$.~\label{c2} 
\end{enumerate}

~\ref{c1} addresses the reality that the model trainer is unlikely to share their local model for intellectual property and privacy
reasons. In spite of the added challenge, we show in~\cref{sec: evaluation} that
the \DOs can effectively share useful data. Additionally, ~\ref{c2} arises from the fact that \DO~will
provide just enough data to convince \MT of their data utility due to privacy and economic concerns.
After utility is established, \MT~may enter into an agreement with the best \DO(s).

%% file: sections/methods.tex
\label{sec:methods}
In this section, we present \method, our data sharing protocol between the \MT and \DOs.



    


\begin{algorithm}
\caption{Mycroft}\label{alg: mycroft_algorithm}
\begin{algorithmic}[1]
    \Require $M_{\MT}$, $D^{\text{test}}$, $\DO$'s loss function $L_{\DO}$, $\DO_{i}$'s dataset $D_i$, Budget $k$,
    \State $\Dhard \leftarrow \test(D^{\text{test}}, M_{\text{MT}})$
    \State $\text{Send $\Dhard$ to the $\DOs$}$
    \State $ \Duseful \leftarrow \DO_{i}\text{ runs } \dataselect$ \algorithmiccomment{\dataselect calls \textbf{OMP} \ref{alg: omp_algorithm} or \textbf{FeatureSimilarity} \ref{alg: featuresimilarity}}
    \State $M_{\text{MT}}' = \train(D^{\MT} \cup \Duseful)$. \text{\hspace{2.28cm} depending on whether $L_{\DO}$ is differentiable}
\end{algorithmic}
\end{algorithm}


\noindent \textbf{Overview of approach:} The overall approach is outlined in
Algorithm~\ref{alg: mycroft_algorithm}. In brief, after each \DO~receives the
dataset $\Dhard$ from \MT, they will use it to identify a subset
$D_i^{\text{useful}}$ of size $k$ from their local training data $D_i$ that is
relevant for \MT to predict $\Dhard$ correctly. To do so, \DO can use either
functional similarity (loss gradient similarity) or feature similarity, or a
combination of both to identify $D_i^{\text{useful}}$, depending on whether the
loss function $L_{\DO}$ is differentiable. The \DO~then sends
$D_i^{\text{useful}}$ to \MT, who will use this data to update their model
$M_{\text{MT}}$. The key technical challenge we address in this section is that
of designing the subroutine \texttt{DataSelect}.

\subsection{Functional Similarity via Loss Gradient Matching}
For models trained using a differentiable loss function, the gradient of the
loss with respect to each model parameter shows how each sample affected the
model during training. Samples that produce similar gradients are considered
functionally similar. This concept has been used in several studies to sub-select
training data to improve the efficiency of model
training \citep{mirzasoleiman2020coresets,killamsetty2021grad}. Ideally, to find
the most relevant samples to $\Dhard$, the model used would be $M_{\text{MT}}$.
However, due to constraint \ref{c1} that no \DO has access to \MT's model, we
assume that each \DO has a model $M_i$ with parameters $\theta_i$ trained using
a differentiable loss function $L$ that can function as a reasonable proxy (our
empirical findings show that this method works well in practice).

We formulate the problem of finding $D_i^{\text{useful}}$ as that of obtaining a
$k$-sparse weight vector $\mathbf{w}$ over $D_i$, with the weight assigned to
each sample corresponding to its utility.
To find this $k$-sparse $\mathbf{w}$, we: (1) find the averaged gradient of the
loss $L$ computed on $\Dhard$ with respect to the parameters $\theta_i$ of $M_i$
(denoted $\nabla_{\theta_i}L(\Dhard)$); (2) compute the gradient of the loss $L$
computed on each sample $z_j \in D_i$ with respect to the parameters
$\theta_i$ of $M_i$; (3) solve the following regularized optimization problem:
\begin{align}%
    \min_{\|\mathbf{w}\|_0 \leq k} e_{\lambda}(\mathbf{w})= \min_{\|\mathbf{w}\|_0 \leq k} \left \lVert \sum_{z_j \in D_i} \mathbf{w}_j \nabla_{\theta_i}L(z_j) - \nabla_{\theta_i}L(\Dhard) \right \rVert + \lambda \left \lVert \mathbf{w} \right \rVert_2^2.
   \label{eq:joint_loss_1}
\end{align}

The first term in Eq.~\ref{eq:joint_loss_1} ensures that a weighted sum of the
selected samples is close to the gradient of the loss on $\Dhard$, while the
regularization term prevents the assignment of very large weights to a single
instance. The $\ell_0$-``norm'' constraint on $\mathbf{w}$ enforces sparsity but
leads to an NP-hard problem. To tackle this issue, we can use a greedy
algorithm, Orthogonal Matching Pursuit (OMP)~\citep{pati1993orthogonal}, to
find a close approximation due to the sub-modularity of
$e_{\lambda}(\mathbf{w})$~\citep{elenberg2016restricted}~\footnote{This use of
OMP is inspired by~\citep{killamsetty2021grad}, who use it for dataset
compression during training.}.  We detail OMP in Algorithm~\ref{alg:
omp_algorithm}. The choice of which state of the model to use (typically stored
as checkpoints) is explored in \cref{subsec: ablation}.


\noindent \textbf{Other potential gradient based techniques:} Before we settled
on utilizing OMP on the model gradients for the \texttt{DataSelect} process, we
also explored various other coreset techniques \citep{zhou2022probabilistic,
coleman2019selection, toneva2018empirical, ducoffe2018adversarial,
ren2018learning}. Since these techniques are designed to create subsets of data
that can approximate the loss of an entire dataset, most of them are not
directly applicable to our setting which requires finding a small subset which
will be useful for a specific task to the $\MT$. Additionally, these techniques
are designed to be used jointly when models are trained from scratch whereas we
are only interested in efficient solutions that can find useful samples by
leveraging an already trained model. Among coreset techniques that can be
adapted to our problem setting, we decide to use OMP because other techniques
either underperform our chosen approach \citep{mirzasoleiman2020coresets} or
are too computationally intensive to be feasible \citep{ren2018learning}.



\subsection{Feature Similarity}

We also find that data samples which are similar to samples from $\Dhard$ in an
appropriate feature space are useful for \MT~to augment their training set. This
is especially useful for models which have been trained \emph{without
differentiable loss functions}. Given any good feature extractor $\phi(\cdot)$
that maps a sample $x_j$ to its feature representation $\phi(x_j)$, we compute
the distances of each sample in $\Dhard$ to each sample in $D_i$ and store them
in a matrix $\Psi$ $\in \mathcal{R}^{|\Dhard| \times |D_{i}|} $. We construct
$\Duseful$ by using a greedy heuristic which first sorts this matrix in the row
dimension followed by the column dimension. We select the top-$k$ samples by
iterating over the columns which selects $|\Duseful|$ samples with the minimum
distance to $\Dhard$ samples while ensuring coverage of $\Dhard$ (see
Algorithm~\ref{alg: featuresimilarity} for more details). The choice of feature
representations $\phi(\cdot)$ and distance function $d(\cdot,\cdot)$ is
contingent on the domain of the data and the classification task. For the image
datasets, we use the feature space of a image retrieval model called Unicom
\citep{an2023unicom} and $L_2$ distance as our distance function. For the
tabular dataset where existing techniques were ineffective, we propose a
\texttt{ExtractBinningFeatures} algorithm using Hamming distances over adaptive
grids (Appendix~\ref{tab:binning_algo}).

\subsection{\texttt{FuncFeat} : Combining Gradient and Feature Similarity}
Whenever the \DO has access to a model trained on $D_i$ as well as a good
feature extractor, they can combine both notions of similarity to find useful
samples. Combining both notions of similarity may improve the quality of
$\Duseful$ as samples which can align both in the feature and gradient space are
more likely to be relevant to $\Dhard$. Our technique (\texttt{FuncFeat})
introduces a regularization term in terms of a composite norm that incorporates
the feature similarity between samples from $\Dhard$ and $D_i$ to the
approximation error from Eq.~\ref{eq:joint_loss_1}: 
\begin{align}
    e'_{\lambda_1,\lambda_2}(\mathbf{w})= e_{\lambda_1}(\mathbf{w}) + \lambda_2 \left \lVert \Psi \mathbf{w} \right \rVert_2^2,
    \label{eq:joint_loss} 
\end{align}
where $\Psi$ is a matrix of distances. The second term functions as a
regularizer that penalizes samples that are far from feature representations of
$\Dhard$. In the following theorem, we show sub-modularity:
\begin{theorem}\label{thm: submodular} If the loss function $L(\cdot)$ is
    bounded above by $L_{\text{max}}$ and $\, \forall j, \, \| \nabla_{\theta_i}
    L (z_j) \| \leq \nabla_{\text{max}}$, then
    $f_{\lambda_1,\lambda_2}(\mathbf{w}) =
    L_{\text{max}}-e'_{\lambda}(\mathbf{w})$ is weakly submodular with parameter
    $\gamma' \geq
    \frac{\lambda_{1}+\lambda_{2}\|\Psi\|_{2}^{2}}{\lambda_{1}+\lambda_{2}\|\Psi\|_{2}^{2}+k
    \nabla_{\max}^{2} }$,
\end{theorem}
where $\|\cdot\|_2$ is the spectral norm for matrices. From
\cite{elenberg2016restricted}, we get that OMP returns a $1-e^{\gamma'}$-close
approximation of the maximum value of $f_{\lambda_1,\lambda_2}(\mathbf{w})$.







%% file: sections/evaluation.tex

In this section, we empirically demonstrate that \method outperforms \randsamp
and rapidly approaches \fullinfo while operating within the communication
constraints laid out in \cref{sec: motivation}. We also present several ablation
studies to investigate key design choices for \method.


\subsection{Experimental Setup} \label{subsec: exp_setup} In this section, we
describe details about the datasets, models, and metrics we use to evaluate
\method, followed by how we construct $\Dhard$ and evaluate \method.

\noindent
\textbf{1. Datasets: } We evaluate \method~
on classification tasks over two domains: computer vision and network traffic
classification. For the computer vision tasks, we use six different datasets,
while for the network traffic classification, we use a tabular dataset that 
represents flow features of network traffic. Further details
are in Appendix~\ref{appsec:dataset_details}. 

\textbf{Image datasets:} 
\begin{itemize}[noitemsep]
  \item \textbf{Food datasets:} We use three datasets from the food computing
  domain: Food-101 \citep{bossard2014food}, UPMC Food-101 \citep{wang2015recipe}
  and ISIA Food-500 \citep{wang2015recipe}. We assign Food-101 to be the $\MT$'s dataset and UPMC Food-101 and ISIA Food-500 as
  datasets of two $\DO$s. 

  \item \textbf{Dog datasets:} We use two datasets for dog breed classification: 
  Imagenet-Dogs \citep{deng2009imagenet} and Tsinghua-Dogs \citep{zou2020new}.
  Imagenet-Dogs contains 120 dog classes from Imagenet whereas Tsinghua-Dogs
  contains 130 dog classes which include all the classes in Imagenet-Dogs. We consider Imagenet-Dogs and Tsinghua-Dogs to be the $\MT$'s and $\DO$'s dataset respectively.
  
 
  \item \textbf{Dogs \& Wolves:} We curate a dataset of Dogs and Wolves with 
  spurious correlations to simulate a controlled $\MT$-$\DO$ interaction, 
  which helps illustrate effectiveness of \method. The $\MT$ model is trained 
  on a dataset containing these spurious correlations, referred to as \dvwS, 
  therefore, the model performs poorly on a dataset without these correlations. 
  We label such a dataset as \dvwN and simulate a $\DO$ which has this dataset. An 
  illustration of this dataset is provided in Figure~\ref{fig:DogsWolves_explanation} in the Appendix.

\end{itemize}

\input{tables/varying_budget.tex}

\textbf{Tabular dataset: } We use the IoT-23 dataset \citep{garcia2020iot23} which
contains tabular features derived from the network traffic flows. The $\MT$'s task is to
select $\DO$s that would improve its model's ability to detect malicious
traffic. In order to make the evaluation comprehensive, we experiment with
different combinations of $\MT$s, $\DO$s, and types of malicious attacks
totalling 665 combinations.

\textbf{2. Models:}
For the image datasets, all our experiments use ResNet50 \citep{he2016deep}
models pre-trained on Imagenet since after evaluating the vision datasets on several CNN architectures, 
including MobileNets, EfficientNets, and ResNets (18 \& 50), we found that our results were consistent across these architectures.
For the tabular dataset, we use Decision Trees, XGBoost and Random Forest as previous 
works \citep{grinsztajn2022tree, hasan2019attack, yang2022efficient} have shown that simple models are not only computationally 
effective but also can outperform deep learning models for typical tabular datasets. Appendix~\ref{appsec:training_details} 
contains details for the training procedure. 

\textbf{3. Metrics \& Baselines:}
We define \fullinfo to be the setting where $\MT$ uses a $\DO$'s
entire dataset to train their model. This represents the upper bound on the
performance improvement with \problem ~from a $\DO$. We use \randsamp as our
baseline technique for \problem~with knowledge of the class label from which
data is to be retrieved. Several works have shown that random sampling is an effective
strategy for dataset compression \citep{mahmud2020survey, guo2022deepcore,
mirzasoleiman2020coresets, killamsetty2021glister}, which is highly relevant to
our task and thus a valid baseline. For the image datasets, we report classification accuracy (normalized between
$0$ and $1$) and F1 score for the tabular dataset (due to class imbalance) for
$\Dhard$.

\textbf{4. Construction of $\Dhard$:} Our approach for creating the $\Dhard$
subset involves two steps. First, we identify the misclassified samples in the
validation set, $D^{\text{val}}$, forming $D^{\text{hard}}$. Then, if
$D^{\text{hard}}$ is large enough, we randomly select a small subset to share
with the $\texttt{DO}$s and keep the remaining for evaluation. If it is too
small to meaningfully split, we share the entire $D^{\text{hard}}$.

\textbf{5. Evaluation setup:} 
For the tabular dataset, where we have enough samples for a meaningful split, we
evaluate the performance of the $\MT$'s model on a separate test dataset not
seen by \MT. In vision tasks, where $\Dhard$ often lacks sufficient samples for
such a split, we evaluate the model directly on the $\Dhard$ dataset. When a
meaningful split is possible on $\Dhard$, we evaluate on the held-out $\Dhard$
data. For evaluation, whenever model gradients are available, we use our \textit{FunctFeat} technique
to construct $\Duseful$. Otherwise, we use our feature similarity methods. 




\subsection{Results}\label{subsec: results} Here, we present the results of
evaluating \method on different datasets and compare it with our \randsamp
baseline. We also vary data budgets to understand how well \method can
approximate the performance of the \fullinfo setting. 


\textbf{Image Datasets:} We present results for the image datasets
Table~\ref{tab:varying_budgets_images} and make three key observations. First,
we can see that \method~outperforms \randsamp across all datasets and at all
budgets of $\Duseful$. Secondly, we note that \method can rapidly converge to
the \fullinfo setting using only a fraction of the dataset. To be specific, the
highest budget on $\Duseful$ amounts to training on at most \textbf{32\%} of the
$\DO$'s dataset, on average, in our experiments, yet \method is able to reach at
least \textbf{83\%} of the \fullinfo performance.

\begin{figure}[t]
  \centering
  \subfloat[CDF of \MT's F1 score after data sharing. As the number of samples in $\Duseful$ increases, both the performance for \randsamp and \method increase and move closer towards \fullinfo performance. \method outperforms \randsamp, as can be seen from the fact that \method with $\Duseful$ of 5 samples outperforms \randsamp with $\Duseful$ of 100 samples.]{
      \includegraphics[width=0.45\textwidth]{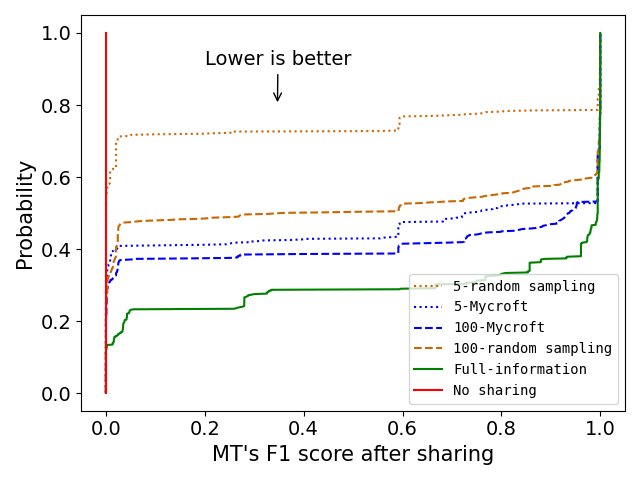}
      \label{fig:cdf_enola_random}
  }
  \hfill
  \subfloat[CDF of $\Duseful$ budget required for \randsamp and \method to match \fullinfo for cases where sharing data is helpful using DecisionTree classifier. For most cases, \method uses a much smaller $\Duseful$ budget compared to \randsamp to match the performance of \fullinfo. N = 474 cases where sharing data is helpful (F1 score for \fullinfo >= 0.5).]{
      \includegraphics[width=0.45\textwidth]{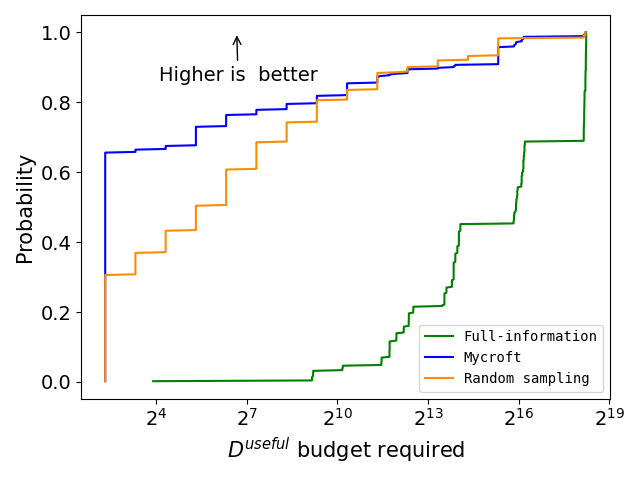}
      \label{fig:histogram_full_enola_score}
  }
  \caption{Performance of \method compared to \randsamp and \fullinfo on the tabular dataset.}
  \vspace{-15pt}
\end{figure}

\textbf{Tabular Dataset:}
Fig~\ref{fig:cdf_enola_random} displays the result for the tabular dataset. We
see that $\Duseful$ budget of 5 samples retrieved using \method outperforms
$\Duseful$ of 100 samples retrieved using \randsamp. In addition, as we increase
$\Duseful$ budget to 100, we move closer to the \fullinfo performance.
To fully explore the performance of \method on various budgets, we start with a
$\Duseful$ budget of 5 samples and double it until the $\MT$'s performance
reaches that of the \fullinfo setting
(Fig~\ref{fig:histogram_full_enola_score}). We observe that \method reaches the
\fullinfo performance using a smaller $\Duseful$ budget compared to
\randsamp, especially on small $\Duseful$ budgets. For example, with a
budget of merely 5 samples, \method can reach the same performance as \fullinfo
on \textbf{65\%} of the combinations used, as opposed to \textbf{30\%} for
\randsamp. Thus, \method significantly reduces the amount of data that needs to
be shared to obtain the same improvement in performance as \fullinfo. For this dataset, we note that \method only uses feature similarity to retrieve
useful data. Further discussion in is \cref{appsec:additional_results}.

\subsection{Ablation studies}\label{subsec: ablation}

We conduct ablation studies on the image datasets to analyze key design choices of \method. For tabular dataset ablations, see Appendix \ref{appsec:tabular_ablation}.

\noindent \textbf{Fine-tuning vs training from scratch:} We compare two
approaches for the $\MT$ to incorporate $\Duseful$ into their model: finetuning
and retraining. We find that retraining from scratch and fine-tuning both
achieve similar performance but fine-tuning is much more computationally
efficient. In particular, training \MT's models with externally augmented data
from scratch takes 8 hours on average on a single Nvidia A40 GPU whereas
finetuning only requires approximately 50 minutes. Hence, we choose to use
finetuning for all our experiments.

\noindent \textbf{Checkpoint selection for loss gradient matching:} We find that
earlier checkpoints provide more useful gradients for gradient matching
(Figure~\ref{fig:omp_ablation}). Please refer to Appendix
\ref{appsec:additional_results} for more details.

\noindent \textbf{Gradient matching with different \MT-\DO model architectures:}
We also explore over the impact on the performance of gradient matching when the
$\DO$ and $\MT$ have different architectures. Concretely, we keep the $\MT$'s
architecture as ResNet50 but change the $\DO$'s architecture to EfficientNetB0
\citep{tan2019efficientnet}. Overall, we observe small differences in the
performance of the augmented model ($<0.01$) but note that gradient matching over different
architectures \citep{jain2024efficient} is an avenue for future exploration.

\noindent \textbf{Other feature spaces for similarity matching:}
Besides Unicom, we also explore other feature spaces, such as that of a pretrained ResNet50
. We evaluate feature spaces on the Dogs \& Wolves dataset and measure how many useful samples the \DO's model
can retrieve from the \dvwN dataset when performing similarity matching in each
feature space. We observe that 76\% of retrieved samples are useful when using
Unicom, whereas only 6\% are useful when using the ResNet50 feature space. We
show a sample of the Top-k retrieved samples in
Figure~\ref{fig:DogsWolves_retrievals}.

\begin{figure}
  \centering
  \includegraphics[width=0.7\textwidth]{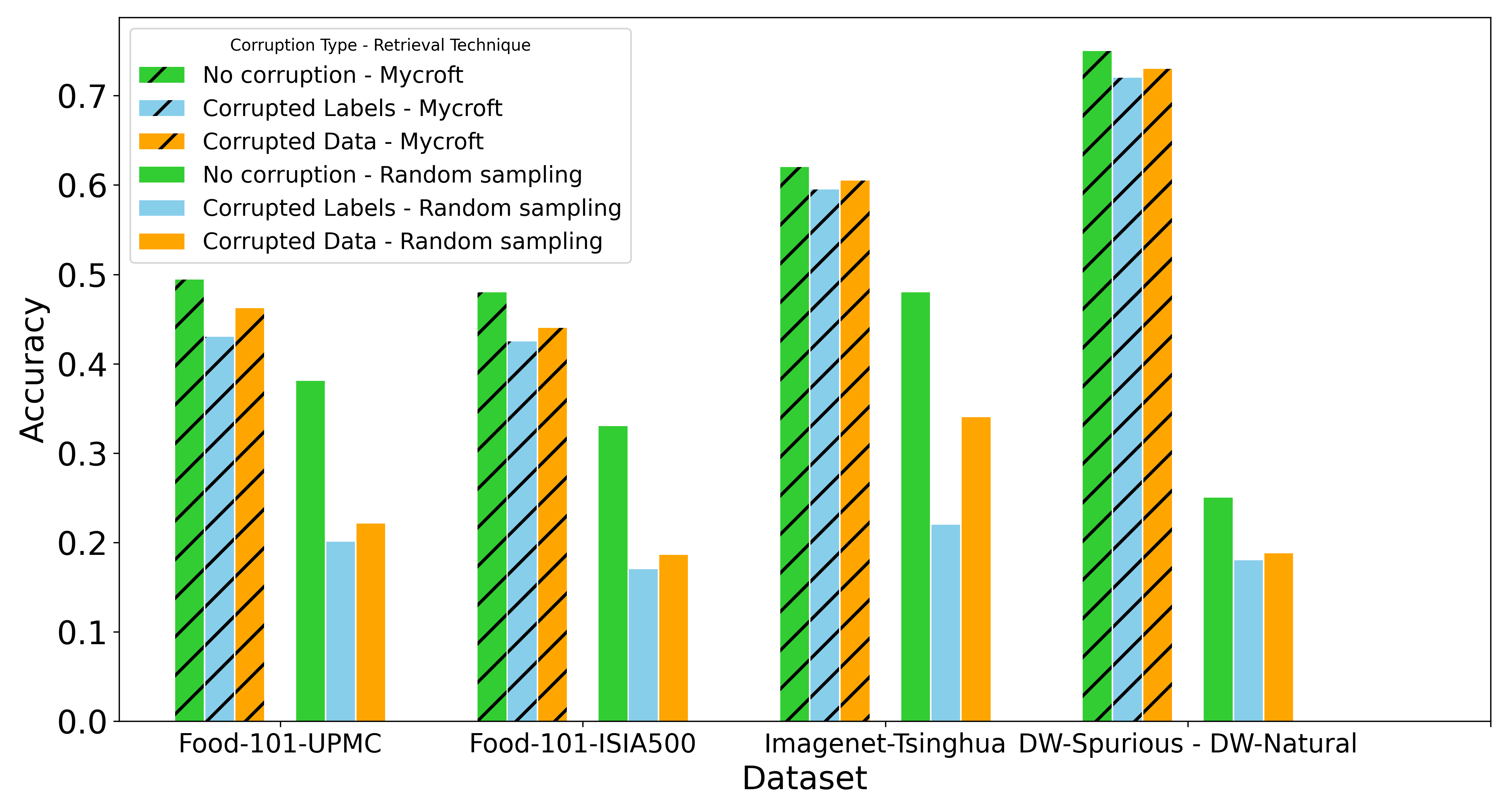}  
  \caption{{Accuracy of $M_{\text{MT}}'$ when trained on $\Duseful$ retrieved using \method and \randsamp under the scenario where approximately 70\% of the data or labels are corrupted.}}\label{fig:scenario:corruptions}
\vspace{-10pt}
\end{figure}


%% file: tables/varying_budget.tex
\begin{table}
  \resizebox{1.0\textwidth}{!}{
  \centering
  \begin{tabular}{crrrrrrrrcc}
    \toprule
     & \multicolumn{2}{c}{\makecell{Food-101 - UPMC}} & \multicolumn{2}{c}{Food-101 - ISIA500} & \multicolumn{2}{c}{Imagenet - Tsinghua} & \multicolumn{2}{c}{\makecell{\dvwS~ -\\ \dvwN}}  \\
    \cmidrule(lr){2-3}\cmidrule(lr){4-5}\cmidrule(lr){6-7}\cmidrule(lr){8-9}
    Budget $k$ & \method & \makecell{\texttt{random}-\\ \texttt{sampling}} &\method & \makecell{\texttt{random}-\\ \texttt{sampling}} &\method & \makecell{\texttt{random}-\\ \texttt{sampling}} & \method & \makecell{\texttt{random}-\\ \texttt{sampling}}  \\
    \midrule
    8 & 0.42 & 0.36  & 0.33 & 0.18 & 0.43 &  0.31 & 0.44 & 0.13 \\
    16 & 0.50 & 0.38 & 0.48 & 0.33 & 0.62 & 0.48 & 0.50 & 0.13 \\
    32 & 0.61 & 0.47 &  0.54 & 0.40 & 0.70 & 0.56  & 0.75 & 0.25  \\
    64 & 0.75 & 0.64 &  0.72 & 0.61 & NA & NA & 0.75 & 0.25 \\
    128 & 0.86 & 0.72 & 0.83  & 0.70 & NA & NA & 0.88 & 0.38 \\
    \bottomrule
  \end{tabular}
  }
  \caption{Accuracy of $M_{\text{MT}}'$ with varying budgets of $\Duseful$.
  \fullinfo results in 100\% \\ accuracy on $\Dhard$. Column headers indicate the $\MT$ and $\DO$ datasets separated by hyphens.
  }\label{tab:varying_budgets_images}
  \vspace{-10pt}
\end{table}

%% file: sections/case_studies.tex
In this section, we examine the applicability of \method across different
real-world data-sharing scenarios. The first three scenarios are evaluated using
datasets.
all our vision datasets, while the last scenario includes results for both the vision and tabular datasets.

\textbf{Scenario 1 - Corrupted data features:}
In real-world scenarios, data features can be corrupted due to factors like
hardware failures as well as collection and transmission errors. To assess
\method's reliability in this context, we corrupt the $\DO$ dataset using random
image transformations, applying random masking, color jitters, etc. The results,
shown in Figure~\ref{fig:scenario:corruptions}, indicate that \method's
performance declines by only 2.7\% on average, compared to a 13.7\% drop for
\randsamp. This demonstrates \method's resilience to corrupted data while still
retrieving useful subsets.


\textbf{Scenario 2 - Corrupted labels:}
Large-scale supervised learning datasets often contain incorrect labels. To
evaluate \method's performance in such conditions, we randomly permuted 70\% of
the labels in \DO's dataset. The results, shown in
Figure~\ref{fig:scenario:corruptions}, demonstrate that \method is highly robust
to label corruption, with an average performance drop of only 4.4\%, compared to
a 16.9\% decrease for \randsamp. This highlights \method's effectiveness in
handling noisy labels.

\textbf{Scenario 3 - \method in missing label settings:} In cases where the $\DO$
dataset lacks labels entirely, \method can rely on feature-space distances. For
vision datasets, \method can use feature distances (Unicom) to retrieve samples
and assign them pseudo-labels by matching each sample to the closest $\Dhard$
sample in the feature space. The results in Table~\ref{tab:enola_no_labels} show
minimal performance degradation between the scenario where labels are available (gradient similarity can be used) and not available (only feature similarity
can be used). 


\input{tables/enola_no_labels.tex}

\textbf{Scenario 4 - Preference ordering for several $\DO$s:}
In data markets, multiple sellers often provide datasets with varying utility. 
In such cases, \method should be able to rank these datasets to support more 
informed data-sharing agreements. We tested this by constructing several 
$\DO$ datasets with different utility levels (details in Appendix 
\ref{appsec:case_studies}) to see if \method could correctly rank them. 
The results, shown in Table~\ref{tab:several_DOs}, 
indicate that \method successfully identifies the most promising datasets, 
while \randsamp struggles to do so.

For tabular data, where clear rankings are harder to establish 
(useful datasets often perform similarly), we measured the number of useful 
$\DOs$ (defined as those achieving an F1 score of 0.5 or higher after data sharing) 
retrieved by \method and \randsamp. As shown in Table~\ref{tab:tabular_results}, 
\method consistently retrieves more useful datasets across different $\MTs$, 
demonstrating its ability to capture the true utility of various datasets. 
Notably, \method sometimes even outperforms \fullinfo in selecting useful data, 
as discussed in Appendix \ref{appsec:case_studies}. 
This highlights \method's suitability for ranking datasets in 
data-sharing scenarios.

\begin{table}
    \parbox{.45\linewidth}{
      \input{tables/several_DOs_DvW.tex} 
    }
    \hfill
    \parbox{.45\linewidth}{
      \input{tables/enola_random_tabular.tex}
      }

    \end{table}

%% file: tables/enola_no_labels.tex
\begin{table}
  \centering
  \begin{tabular}{crrrccc}
    \toprule
    $\MT$ - $\DO$ & {\makecell{\method \\ No labels}} & \makecell{\texttt{random}\\ - \texttt{sampling}  \\ No  labels} & {\makecell{\method  \\ With labels}} \\
    \midrule
    \texttt{Food-101 - UPMC} & 0.58 & 0.19 &  0.61\\
    \texttt{Food-101 - ISIA500} & 0.47 & 0.14 & 0.54 \\
    \texttt{Imagenet - Tsinghua} & 0.44 & 0.25 & 0.7 \\
    \texttt{DvW - Spurious - DvW - Natural} & 0.68 & 0.21 & 0.75  \\  
    \bottomrule
  \end{tabular}
  \caption{
  Accuracy of \method compared to \randsamp for \textbf{Scenario 3} where the $\DO$ has an unlabelled dataset. In this case, \method utilizes feature 
  similarity to construct $\Duseful$. We compare the accuracy in this scenario with the performance of \method when the $\DO$ has labels and utilizes
  gradient similarity.
   }\label{tab:enola_no_labels}
\end{table}

%% file: tables/several_DOs_DvW.tex
\resizebox{.45\textwidth}{!}{
  \centering
  \begin{tabular}{crrrccc}
    \toprule
    $\DO$ & {\makecell{\method}} & \makecell{\texttt{random}-\\ \texttt{sampling}} & \makecell{\texttt{full}-\\ \texttt{information}} \\
    \midrule
    \texttt{DO-1} & 0.81 & 0.18 & 0.88 \\
    \texttt{DO-2} & 0.63 & 0.31 & 0.69 \\
    \texttt{DO-3} & 0.44 & 0.25 & 0.63   \\
    \texttt{DO-4} & 0.56 & 0.25 & 0.50  \\
    \texttt{DO-5} & 0.19 & 0.25 & 0.13 \\

    \bottomrule
  \end{tabular}
  }
\caption{Preference ordering (based on accuracy) generated from \method~and \randsamp for selecting from among several $\DO$ candidates with different levels of utility (where each $\DO$'s number corresponds to their utillity) from \textbf{Scenario 4}. 
\method is mostly able to retrieve the ground-truth preference ordering whereas \randsamp fails to do so.}\label{tab:several_DOs}

%% file: tables/enola_random_tabular.tex
\resizebox{.45\textwidth}{!}{

    \centering
    \begin{tabular}{crrr}
    \toprule
    \MT & {\makecell{\method}} & \makecell{\texttt{random}-\\ \texttt{sampling}} & \makecell{\texttt{full}-\\ \texttt{information}} \\
    \midrule
    \texttt{MT-1} & 49& 34& 80  \\
    \texttt{MT-2} & 16& 5& 40  \\
    \texttt{MT-3} & 76& 58& 92  \\
    \texttt{MT-4} & 87& 23& 90  \\
    \texttt{MT-5} & 60& 37& 60  \\
    \texttt{MT-6} & 0& 0& 24  \\
    \texttt{MT-7} & 92& 25& 88  \\
    \bottomrule    
    \end{tabular}
}
    \caption{Number of useful \DOs (F1 score of $M_{\text{MT}}'$>=0.5) retrieved by \method and \randsamp for budget of 5 samples and by \fullinfo for different \MTs for tabular data. Number of \DO candidates = 95.}\label{tab:tabular_results}

%% file: sections/relatedworks.tex
\textbf{Data augmentation and synthetic data generation:} Numerous studies have
investigated ways to enhance training data quality by leveraging existing
datasets through methods such as image overlay~\citep{inoue2018data}, random
erasure~\citep{zhong2020random}, and common data augmentation techniques like
rotation and cropping~\citep{inoue2018data,zhong2020random,chlap2021review,shorten2019survey,hussain2017differential,feng2021survey}.
These strategies are designed to generate new samples, reduce overfitting, and
improve model generalization. However, they are less effective if the test data
deviates significantly from or is underrepresented in the training set.
Similarly, using generative models to create additional training samples~\citep{tripathi2019learning, such2020generative, 10.1145/3652963.3655071} can
enhance performance, but this approach often requires substantial data to train
the generative model itself. Moreover, synthetic data may fail to capture the
full complexity of real-world data distributions, limiting its impact on hard
sample performance.

 \textbf{Data augmentation using publicly available data:} Researchers have also
 explored sourcing data from publicly available data lakes~\citep{9693372,
 10.14778/3476311.3476346, 10.1145/2213836.2213848, Esmailoghli2021COCOACC,
 10.1145/3448016.3458456, 10.14778/2994509.2994534, 8731486, aurum,
 galhotra2023metam}. For example, METAM~\citep{galhotra2023metam} profiles
 various datasets to identify and select the most relevant ones, thereby
 improving the quality of training data and enhancing performance on downstream
 tasks. Another approach, Internet Explorer~\citep{li2023internet}, retrieves
 task-specific images from the internet to support self-supervised learning.
 However, these solutions rely on open access to data and are ineffective in
 scenarios where data access is restricted. Consequently, they do
 not address the challenge of sourcing data from private entities.

 \textbf{Selecting data without full information:} Recent work called
Projektor~\citep{kang2024performance} addresses the problem of selecting and
weighting useful data owners without full information about the underlying data.
Projektor assumes that some of the \DO’s data is publicly available as pilot
data and can predict the usefulness of the entire dataset based on this subset.
This assumption may not be true in our context as the publicly available pilot
data is not selected based on any knowledge about the \MT’s task, and may be
irrelevant for \MT. Techniques like Projektor can complement \method by finding
the best combination of relevant samples from different \DOs.
 

\textbf{Deriving coresets for large datasets:} There is also extensive research
 on identifying useful subsets of datasets, such as work on ``coresets"
~\citep{killamsetty2021retrieve,kim2022defense,guo2022deepcore}, which focus on
 selecting a representative subset that approximates the cost function of the
 entire dataset. However, most of these methods are better suited for dataset
 compression rather than creating task-specific subsets. In our context,
 approximating the entire cost function is irrelevant to the model trainer's
 needs. Instead, we prioritize selecting a subset most relevant to the specific
 task and have adapted some of these techniques accordingly.

%% file: sections/discussion.tex
\noindent \textbf{Why does \method help?} The fact that \method outperforms
\randsamp in almost all scenarios and quickly approaches \fullinfo, even when
operating under a limited data budget has scenario-dependent explanations.
First, for many \DOs, the majority of their data is irrelevant to the $\MT$'s
task, as such, \randsamp is inefficient and sharing all their data is
unnecessary. Secondly, in many cases, \MT may only need a small subset of the
data to drastically improve their model's performance. For example, in the Dogs
\& Wolves dataset, only a small subset of the data is needed break the spurious
correlations and improve generalization. Similarly, for the IoT attack data, the
significant distinction between attack distributions implies a small subset is
sufficient to differentiate between benign and malicious traffic, and different
kinds of malicious traffic.

\noindent \textbf{Limitations and Future Work:} One limitation of our work is
that it assumes that the model trainer already has access to a set $\Dhard$
that they want to improve on. This may often be a reasonable assumption because
model trainers are looking to improve on the predictions of $\Dhard$ that they
perform poorly on or have low confidence about. However, the case where $\Dhard$
is too limited and therefore, does not give the \dataowners enough of a signal
to determine useful data, is an important for future work. In addition, our method also assumes that the most useful data to share are
likely those that are similar to $\Dhard$. In reality, data that are not close
to $\Dhard$ might also be useful to share. In this paper, we do leverage
functional similarity via gradient matching to include some diverse data points.
However, explicitly promoting diverse data selection is a key direction for
future work. 

Finally, there remains the issue that model trainers may not be willing to share
even small subsets of their data with the data owners, in which case more
privacy-preserving methods can be explored on top of \method. Future directions
to address the privacy limitations include developing a version that avoids
directly sharing data samples but still provides useful information to the
\modeltrainer, such as using techniques like noise addition, feature (instead of
data) sharing, synthetic data generation, and other privacy-preserving methods
on top of \method.
Additionally, exploring versions of \method that are resistant to strategic, and
even malicious, \dataowners is a promising area for future work. 

We hope our approach in this paper lays the foundation for more efficient and
privacy-focused data-sharing frameworks, ultimately democratizing the training
of highly performant ML models.



%% file: sections/appendix.tex
In this Appendix, we aim to (i) provide proofs for the optimality of \method (ii)
provide more details about the datasets, algorithms and models used to obtain the
results in the main body of the paper (iii) present additional experiments to
further validate the discoveries in the main body of the paper. The Appendix is
organized as follows:
\begin{enumerate}
  \item Summary of symbols and notations used (\cref{appsec:symbols})
  \item Proof of the optimality of \method (\cref{appsec:proofs})
  \item Runtime analysis and algorithms for \method subroutines (\cref{appsec:runtime_analysis})
  \item Further details about the experiment setup including the datasets and models used (\cref{appsec:dataset_details})
  \item Additional results and ablation studies (\cref{appsec:additional_results})
  \item Further details and results for case studies (\cref{appsec:case_studies})
\end{enumerate}

\section{Symbols and Notations}\label{appsec:symbols}
\begin{table}[ht]
   \centering
  \begin{tabular}{ll}
  \toprule
  Symbol & Description \\
  \midrule
  $\DO$ & Data Owner \\
  \midrule
  $\MT$ & Model Trainer \\
  \midrule
  $D^{\text{MT}}$ & $\MT$'s dataset \\
  \midrule
  $M_{\text{MT}}$ & $\MT$'s model \\
  \midrule
  $D^{\text{test}}$ & $\MT$'s test dataset \\
  \midrule
  $D_i$ & $i^{th}$ $\DO$'s dataset   \\
  \midrule
  $\Duseful$ & Subset of $D_i$ retrieved by \method or \randsamp   \\
  \midrule
  $\Dhard$ & Subset of $D^{\text{test}}$ which is incorrectly classified and is shared with the $\DO$s \\
  \bottomrule
  \end{tabular}    
  \caption{Table of notations used in the paper.} \label{tab:notations}
  \vspace*{-0.6cm}

\end{table}
  
\section{Proofs}\label{appsec:proofs} In this section, we prove the weak
submodularity of the function obtained by subtracting the error function in Eq.
\ref{eq:joint_loss} from the maximum value of the loss. That is, we need to prove
that the following function is weakly submodular with paramter $\gamma'$:
\begin{align}
  f_{\lambda_1,\lambda_2}(\mathbf{w})= L_{\text{max}} - \min_{\|\mathbf{w}\|_0 \leq k} \left \lVert \sum_{z_j \in D_i} \mathbf{w}_j \nabla_{\theta_i}L(z_j) - \nabla_{\theta_i}L(\Dhard) \right \rVert + \lambda_1 \left \lVert \mathbf{w} \right \rVert_2^2 + \lambda_2 \left \lVert \Psi \mathbf{w} \right \rVert_2^2,
  \label{eq:joint_loss} 
\end{align}
under the conditions specified in Theorem~\ref{thm: submodular}. Given that this
function is submodular, then the use of the Orthogonal Matching Pursuit (OMP)
algorithm from \citet{elenberg2016restricted} will return a $k$-sparse subset
with performance that is a $1-e^{\lambda'}$ approximation of the maximum value.

\begin{proof}[Proof of Theorem~\ref{thm: submodular}]
  From \citet{elenberg2016restricted}, a function is $\gamma'$ weakly submodular
  with $\gamma' \geq \frac{m}{M}$ where $m$ is the restricted strong concavity
  parameter and $M$ is the restricted smoothness parameter. 

  To prove that $f_{\lambda_1,\lambda_2}(\mathbf{w})$ is strongly concave with
  parameter $m$, we need to show that 
  \begin{align}
    -\frac{m}{2} \| \mathbf{v} -\mathbf{w} \|_2^2 \geq  f_{\lambda_1,\lambda_2}(\mathbf{v}) - f_{\lambda_1,\lambda_2}(\mathbf{w}) - \langle \nabla f_{\lambda_1,\lambda_2}(\mathbf{w}), \mathbf{v} - \mathbf{w} \rangle
  \end{align}

  Plugging in $f_{\lambda_1,\lambda_2}(\cdot)$ from Eq.~\ref{eq:joint_loss_1}, we get
  \begin{align*}
    -\frac{m}{2} \| \mathbf{v} -\mathbf{w} \|_2^2 &\geq -\lambda_1  \| \mathbf{v} - \mathbf{w} \|_2^2  - \lambda_2 \| \Psi \mathbf{v} - \Psi \mathbf{w} \|_2^2 \\
    & \geq  -\lambda_1  \| \mathbf{v} - \mathbf{w} \|_2^2 - \lambda_2 \|\Psi \|_2^2 \| \mathbf{v} -\mathbf{w} \|_2^2,
  \end{align*}
where the final inequality arises from the property of the induced norm with
respect to a matrix and $\| \Psi \|$ is the spectral norm of the distance matrix
$\Psi$. This implies $m \leq 2(\lambda_1 + \lambda_2 \|\Psi \|_2^2$).

To prove that $f_{\lambda_1,\lambda_2}(\mathbf{w})$ is restricted smooth with
parameter $M$, we need to show that 
\begin{align}
   f_{\lambda_1,\lambda_2}(\mathbf{v}) - f_{\lambda_1,\lambda_2}(\mathbf{w}) - \langle \nabla f_{\lambda_1,\lambda_2}(\mathbf{w}), \mathbf{v} - \mathbf{w} \rangle \geq -\frac{M}{2} \| \mathbf{v} -\mathbf{w} \|_2^2
\end{align}
Expanding the term on the L.H.S. again, we get, 
\begin{align*}
  &-\lambda_1  \| \mathbf{v} - \mathbf{w} \|_2^2 - \lambda_2 \|\Psi \|_2^2 \| \mathbf{v} -\mathbf{w} \|_2^2 - \sum_j \mathbf{v}_j (\sum_k (\mathbf{w}_k - \mathbf{v}_j) \nabla_{\theta_i}(z_j)^\intercal \nabla_{\theta_i}(z_k)) \\
  & \geq  -\lambda_1  \| \mathbf{v} - \mathbf{w} \|_2^2 - \lambda_2 \|\Psi \|_2^2 \| \mathbf{v} -\mathbf{w} \|_2^2 - k \nabla_{\text{max}}^2 \| \mathbf{v} - \mathbf{w} \|_2^2,
\end{align*}
where the final inequality arises from the $k-$ sparse condition on the weight
vectors and the bound on the gradients of the loss function. This gives $M \geq 2(\lambda_1+\lambda_2 \| \Psi \|_2^2 + k \nabla_{\text{max}}^2)$.

Together, this gives $\gamma' \geq \frac{\lambda_1+\lambda_2 \| \Psi \|_2^2}{\lambda_1+\lambda_2 \| \Psi \|_2^2 + k \nabla_{\text{max}}^2}$.

\end{proof}

\section{Pseudo code and runtime analysis for \method}\label{appsec:runtime_analysis}
\subsection{Pseudo code for \method}




\begin{algorithm}
  \caption{OMP}\label{alg: omp_algorithm}
  \begin{algorithmic}[1]
      \Require $\Dhard$, $\DO$'s loss function : $L$, $D_{i}$, $M_{i}$'s parameteres $\theta$, regularization coefficients: $\lambda_{1}$,$\lambda_{2}$   , subset size: $k$, tolerance: $\epsilon$
      \State $\mathcal{X} \leftarrow \emptyset$ 
      \State $r \leftarrow \nabla_{\theta_i}L(\Dhard) $
  
      \While{$\mathcal{X} \leq k$ and $r$ $\geq$ $\epsilon$ }
      \State $m \leftarrow \argmax_j |Proj(\nabla_{\theta_i}L(D_{i}), r)|$
      \State $\mathcal{X} \leftarrow \mathcal{X} \cup \{m\}$
      \State $w^{*} \leftarrow \argmin_w e'_{\lambda_1,\lambda_2}(w, \mathcal{X})$
      \State $r \leftarrow r - Proj(\mathcal{X}, w^{*})$
      \EndWhile
  
    \State \Return$\mathcal{X}$, w
  \end{algorithmic}
  \end{algorithm}

\begin{algorithm}
\caption{FeatureSimilarity}\label{alg: featuresimilarity}
\begin{algorithmic}[1]
    \Require $\Dhard$, $D_i$, Budget $k$, 
    \State  $ \phi(\Dhard), \phi(\Duseful)  \leftarrow \DO \text{ runs } \textbf{FeatureExtractor}(\Dhard, D_i)$  \algorithmiccomment{\textbf{Unicom} or \textbf{Binnning}}
    \State $ \Psi \leftarrow \textbf{ComputeDistances}(\phi(\Dhard), \phi(\Duseful))$ 
    \State $ \Duseful \leftarrow \textbf{RetrieveTopK}(\Psi, k)$
    \State \Return $ \Duseful$
\end{algorithmic}
\end{algorithm}






\begin{algorithm}
\caption{BinningDistance}\label{tab:binning_algo}
\begin{algorithmic}[1]
    \Require $\Dhard$, $\DO$, percentage to sample from $\DO$: $r$, minimum number of non-empty bins for each feature: $b$, binning candidates list (in increasing order): $candidates$, features used for binning: $features$
    \State $\mathcal{X}^{\DO}$, $\mathcal{X}^{\Dhard} \leftarrow \text{ExtractBinningFeatures}(\Dhard, \DO,r,b, candidates, features  )  $ 
    
    \State $D \leftarrow \emptyset$

    \For{$p^{\DO}$ in $\mathcal{X}^{\DO}$}
    \State $d\_l \leftarrow \emptyset$
    \For {$p^{\Dhard}$ in $\mathcal{X}^{\Dhard}$}
    \State $d \leftarrow \text{GetDistance}(p^{\DO}, p^{\Dhard})$
    \State $d\_l \leftarrow d\_l \cup \{d\}$

    \EndFor
    \State $D \leftarrow D \cup \{d\_l\}$

    \EndFor

    \State \Return~$D$

\end{algorithmic}\label{tab:binning_algo}
\end{algorithm}


\begin{algorithm}
\caption{ExtractBinningFeatures}
\begin{algorithmic}[1]

    \Require $\Dhard$, $\DO$, percentage to sample from $\DO$: $r$, minimum number of non-empty bins for each feature: $b$, binning candidates list (in increasing order): $candidates$, features used for binning: $features$
    \State $\DO^{\text{samples}}$ $\leftarrow$ \text{sample}($\DO$,  $r$)
    \State $ \text{UnionSamples} \leftarrow \Dhard \cup \DO^{\text{samples}}$
    
    \State $\mathcal{X}^{\DO} \leftarrow \emptyset$
    \State $\mathcal{X}^{\Dhard} \leftarrow \emptyset$
    
    \For{$f$ in $features$}
     
    \State $\text{NumBin} \leftarrow \max({candidates})$
    \For {$n$ in $candidates$}
    \State $\text{NumFilled} \leftarrow \text{CountNonEmptyBins}(\text{UnionSamples}[f], n)$

    \If{$\text{NumFilled}\geq b$}
    \State $\text{NumBin} \leftarrow b$
    \State break
    \EndIf
    \EndFor
    \State $\text{edge}^{f} \leftarrow \text{GetEdge}(\text{UnionSamples}[f], \text{NumBin})$
    \State $\mathcal{X}^{\DO} \leftarrow \mathcal{X}^{\DO} \cup \text{GetBinningCoordinates}(\DO[f], \text{edge}^{f})$
    \State $\mathcal{X}^{\Dhard} \leftarrow \mathcal{X}^{\Dhard} \cup \text{GetBinningCoordinates}(\Dhard[f], \text{edge}^{f})$
    \EndFor
    \State \Return $\mathcal{X}^{\DO}$, $\mathcal{X}^{\Dhard}$

  \end{algorithmic}
\end{algorithm}

\subsection{Complexity \& Runtime of \method}\label{runtime_analysis}
\textbf{Image datasets:}
\method~for the image domain consists of two techniques: Unicom and GradMatch. Here, we discuss the computation and memory complexity of both these techniques in order
to give a sense of their efficiency and practicality. Unicom operates by projecting all data points in the representation space of the Unicom model, which is based on CLIP~\cite{radford2021learning},
and computing distances between those data points. Thus, its compute and memory requirement scale in proportion to the number of data points to be projected as each sample requires a forward pass
through the model to acquire its feature representation which needs to be held in stored for computing distances with other data points. Empricially, we find that this procedure takes less than 5 minutes and 
requires less than 4 GB of GPU memory for each experiment we present in this paper. 
GradMatch requires computing gradients for each data point in $\Dhard$ and  $\DO$'s using the $\DO$'s model once. In practice, we only use the gradients of the last two layers, which significantly reduces the compute
and memory requirements. The gradients are then used to run the OMP algorithm which has a complexity of $\mathcal{O}(NM + Mk + k^3)$ for each of the $k$ iterations where $k$ is $|\Duseful|$, $M$ is the dimension of the gradients and $N$ is $|D_i|$. For experiments in this paper, each experiment ran in under 10 minutes and required approximately 6 GigaBytes of memory.

\textbf{Tabular dataset:}
The runtime for tabular dataset as described in \ref{tab:binning_algo} is $\mathcal{O}(MNF)$ where $M$ is the number of samples in $\DO$, $N$ is the number of samples in $\Dhard$, $F$ is the number of features used for \textit{ExtractBinningFeatures}. Empirically, each experiment takes less than 5 minutes and requires less than 3 GB of memory.

\section{Further setup details}\label{appsec:dataset_details}

\begin{figure}[ht]
  \centering
  \fbox{\includegraphics[width=\linewidth]{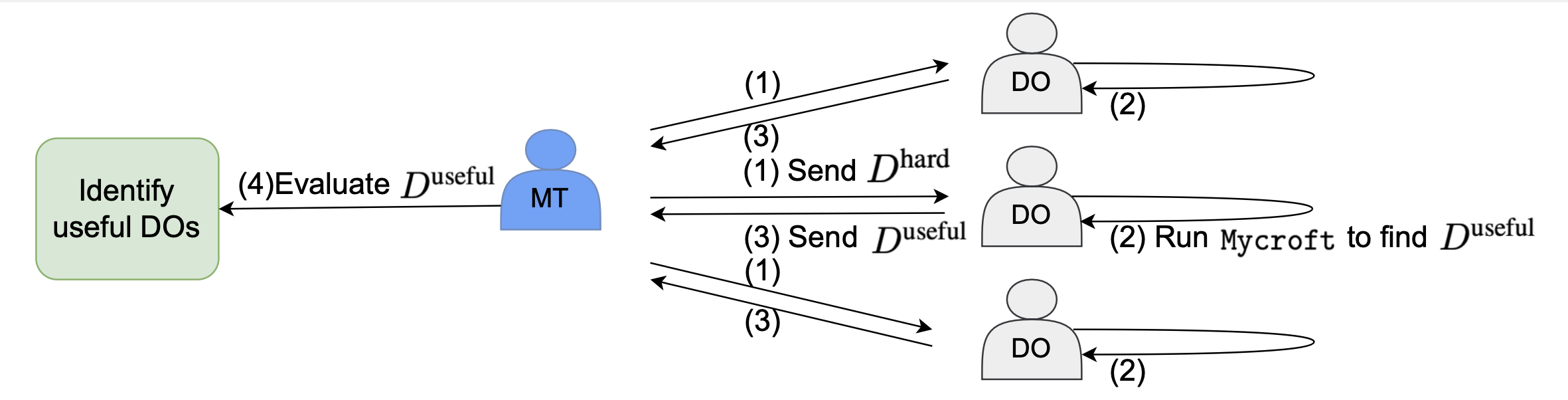}}
  \caption{Framework for \MT~to evaluate the utility of \DO's data.}\label{fig:MT_DO_framework}
\end{figure}

\MT will evaluate the utility of \DO's data based on the framework found in Figure \ref{fig:MT_DO_framework}. Further details about the datasets and training process used to obtain the results in the main body of the paper are provided in this section.

\subsection{Datasets}
\subsubsection{Dogs \& Wolves dataset}\label{appendix:Dogs_Wolves_detail}
\begin{figure}[t]
  \centering
  \captionsetup{justification=centering}
  \includegraphics[width=6cm]{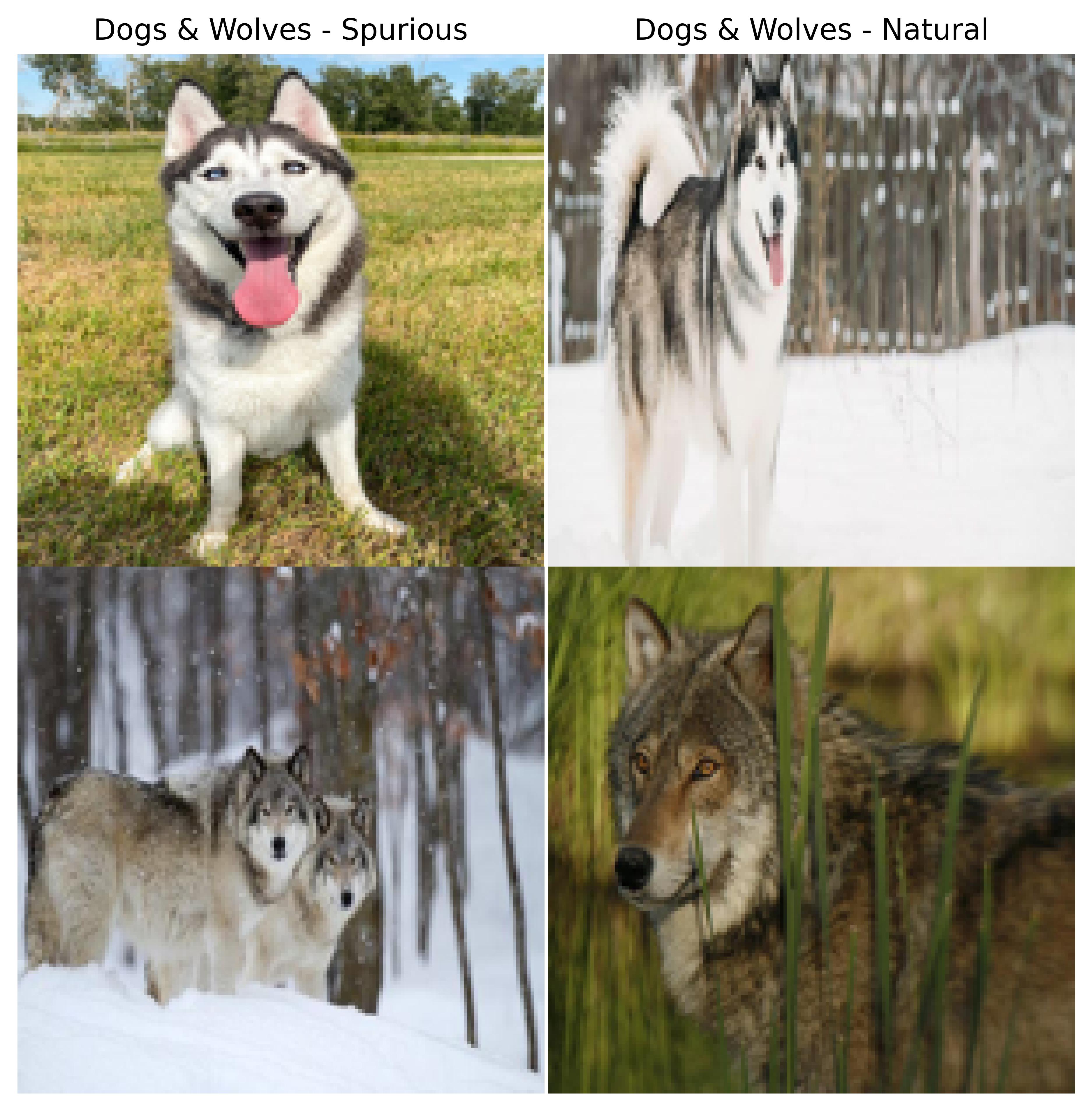}
  \caption{Subsets in the Dogs \& Wolves dataset. The first column shows \dvwS where the dogs
  are on a grass background and the wolves are on snow. The second column shows \dvwN where the dogs are on snow and the wolves are on grass.}\label{fig:DogsWolves_explanation} 
\end{figure}

Neural networks are known to learn spurious correlations in supervised settings. While test data containing the correlations learnt during training often gets classified correctly, data
which does not contain such correlations is prone to misclassification. We exploit this phenomenon to create a dataset which helps us simulate a controlled $\MT$-$\DO$ interaction. In our case, the $\MT$ has a training and
validation dataset which contains spurious correlations but a test dataset which does not contain them and thus their model suffers on the test dataset. \\ Concretely, we curate a dataset which consists of two classes: Dogs and Wolves.
Spurious correlations are introduced in it by controlling the background of each image which can either be snow or grass.
The \MT~has data from both animals being on one type of background. In particular, the dogs are on grass and the wolves are on snow. In the absense of negative examples, the model takes
a shortcut by associating the true label with the background and not the animal. We refer to $\MT$'s training and validation subset as \dvwS. However, the model performs poorly when
the test samples do not contain the spurious correlations i.e., dogs on snow and wolves on grass. We refer to this subset as \dvwN. We simulate a $\DO$ which has a dataset containing both
\dvwS and \dvwN and thus their model does not learn background related spurious correlations. An illustration of this dataset is provided in Figure~\ref{fig:DogsWolves_explanation}.\\ 
Now, if the $\MT$ wants to perform well on data from \dvwN,
they must acquire data from that distribution and retrain their model in order to break the spurious correlations. This motivates the $\MT$ to acquire new data form the $\DO$. It should be noted that the $\MT$ is oblivious
as to why their model performs poorly on \dvwN.

\begin{figure}
\includegraphics[width=\linewidth]{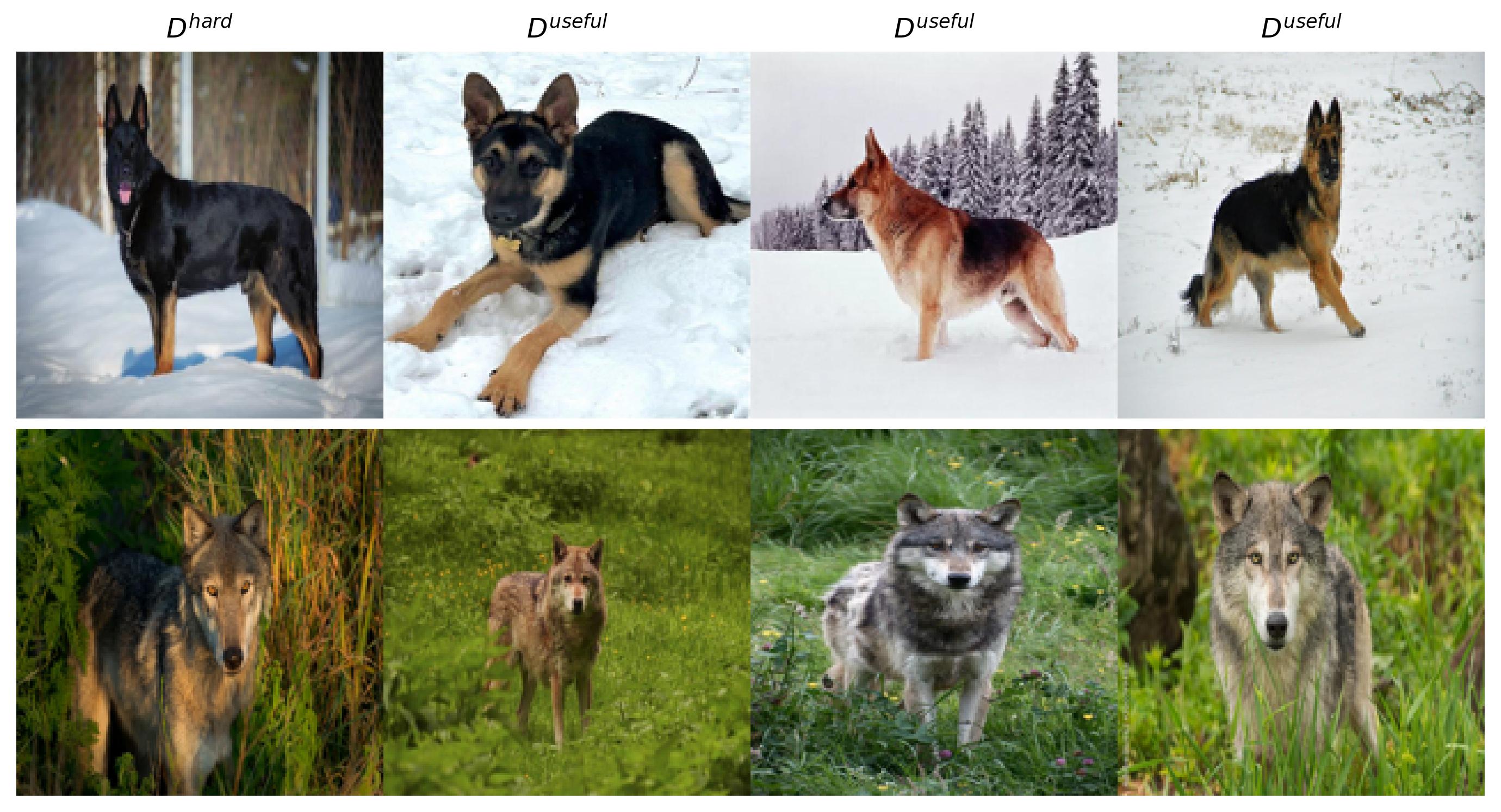}
\caption{Top-k retrieved $\Duseful$ samples using Unicom for $\Dhard$ from the Dogs \& Wolves dataset.}\label{fig:DogsWolves_retrievals}
\end{figure}

\subsubsection{Food dataset}
Food-101 contains 101,000 images of 101 food classes with 750 and 250 images for each category for training and testing. UPMC Food-101 is a twin dataset to Food-101 and thus has the same number of
categories and size as Food-101. ISIA Food-500 is another food recognition dataset with approximately 400,000 images for 500 food classes and has many classes which intersect with the set of classes in UPMC Food-101. 
In our experiments, we use Food-101 as the $\MT$'s dataset and UPMC Food-101 and ISIA Food-500 as datasets of two different $\DO$'s.

\subsubsection{Tabular dataset}\label{appsec:iot_dataset}

\textbf{Data source:}\\
The dataset we use consists of five captures (scenarios) of different IoT network traffic \cite{garcia2020iot23}. Each network consists of traffic of two types : benign traffic (when the IoT devices are not not under attack) and malicious traffic (when the devices are under attack).
The attacks are executed in a Raspberry Pi and each capture can suffer from different attacks.
Details about the types of benign/attack present in each capture can be found in Table \ref{tab:capture_details}.

\begin{table}[h]
    \caption{Attacks present in each capture, Benign stands for Benign traffic. PHP stands for Part Of A Horizontal PortScan attack, CC stands for C\&C attack, CCT stands for C\&C Torii.}\label{tab:capture_details}
    \centering
    \begin{tabular}{lll}
    \toprule
        Capture& Benign/Attacks present \\ \midrule
        1 & Benign, PHP, \\ 
        3 & Benign, CC, PHP \\ 
        20 &  Benign, CCT\\ 
        21 & Benign, CCT\\ 
        34 & Benign, CC, PHP\\
        \bottomrule
    
    \end{tabular}
    
\end{table}

\textbf{Data Labeling:}\\
From the raw pcap files (which contain the raw network traffic data), we use the Python library NFStream \cite{NFStream} to extract feature flows (in tabular format) from pcap files. We then match the timing of the flow with the timing that the attack was executed as mentioned in the data source \cite{garcia2020iot23} to label the data.
After labelling the flows, we split these flows based on whether they are benign or malicious based on their attack type. 

\noindent \textbf{Model Trainers:} 
We define an \MT to be the IoT device in the captures that want to improve its model's ability to predict some particular attack. In this study, we have 7 different \MTs.
In addition, because the attack data in this dataset has quite uniform distribution (most likely because they are conducted in a lab-setting) such that if the model has been trained on the attack, they are very likely to predict a future attack of the same type with high accuracy, 
we assume that these \MTs are only trained on benign data. In reality, this scenario is possible because if the network is new, the chances of them being trained on attack data for this new network is low. 
The \MTs see a very small number of attack data which its model fail to predict and would like to get more data from \DOs~to improve their models.

\noindent \textbf{Data Owners:} We artificially inflate the number and complexity of \DOs~by mixing data from different captures to generate 95 new \DOs. 
This will increase the difficulty of finding relevant samples in \DOs~and simulate the real scenarios where \DO~are often quite complex.

\noindent \textbf{Details for $\Dhard$:}The quantity of $\Dhard$ which each \MT possesses is very small (2\% of the malicious data) and is chosen randomly from the malicious data of the \MT's dataset.

\subsection{Training details}\label{appsec:training_details}
Here, we provide more details about the training procedure we used for obtaining the neural networks we use for our computer vision tasks. \\

For the public image datasets we use, we observe that the model performs well on most classes perform and thus, there is little to gain from \problem. Therefore, to simulate a more realistic setting, we reduce performance on certain classes of
the $\MT$'s model by training them with limited training data. On average, we use 10\% of the original training data for the classes we choose to augment using external datasets 
and attain an average accuracy of 65\% for them. 

The $\MT$'s and $\DO$'s models are ResNet50 models pretrained on Imagenet and finetuned on their respective datasets. The $\MT$'s and $\DO$'s base model is trained for 120 epochs using a learning rate of 0.03 with a cosine annealing weight decay.
The $\MT$'s augmented models, $M_{\text{MT}}'$, are obtained by finetuning their base models for 25 epochs on $\Duseful$ and their original training dataset.


\section{Additional results \& Ablation studies}\label{appsec:additional_results}
Here, we evaluate which training phase of the $\DO$'s model provides the most useful gradient information for data selection. For evaluation, we chose the Dogs \& Wolves dataset as the $\DO$ 
and perform gradient matching for several checkpoints. We compare the percentage of samples in the retrieved subset which belong to the \dvwN data subset since those are the only useful
samples in the $\DO$'s dataset. We plot the results in in Figure~\ref{fig:omp_ablation} and observe that earlier checkpoints indeed provide more useful gradients.

\subsection{Image datasets}\label{appsec:image_ablation}

\begin{figure}
  \centering
  \includegraphics[width=0.5\textwidth]{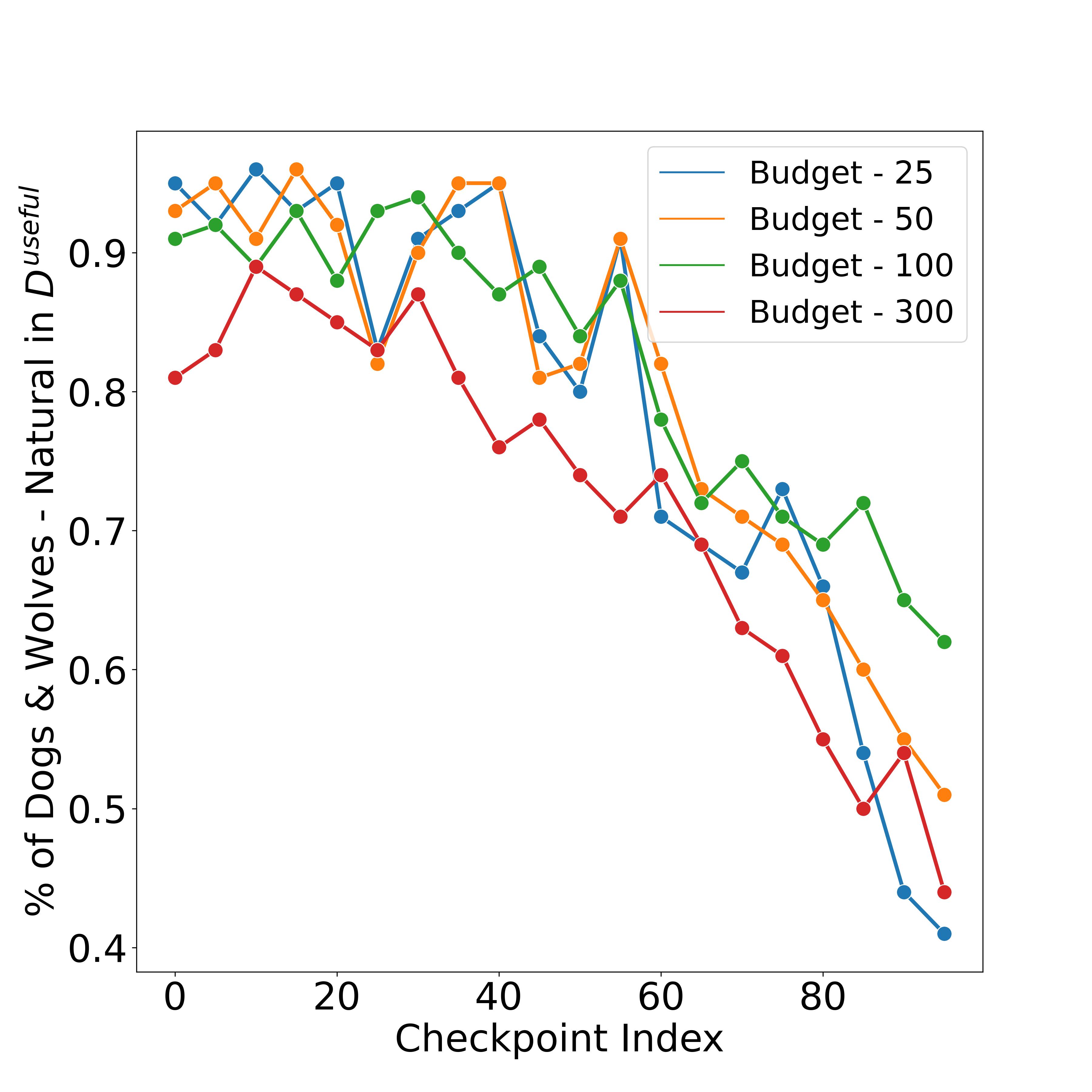}  
     \caption{Effect of the checkpoint used for GradMatch on $\Duseful$. Earlier checkpoints tend to provide more useful gradient information
      for the OMP algorithm.}
     \label{fig:omp_ablation}
\end{figure}

We conduct ablation study on image datasets to see how the selection of checkpoints affect GradMatch's ability to select $\Duseful$. We find that earlier checkpoints tend to provide more useful gradient information for the OMP algorithm. Refer to Figure \ref{fig:omp_ablation} for the results.

\subsection{Tabular data}\label{appsec:tabular_ablation}

For this dataset, as noted in \cref{subsec: results}, the results presented only use
feature similarity. We discuss the reasons here. First, the models that perform
the best on this dataset are tree-based classifiers for which gradient matching
does not apply. In addition, to resemble the effect of gradient matching for
tabular data, we have also attempted to retrieve $\Duseful$ based on \DO's model
confidence score and DecisionTree's decision path when trained on \DO's data. We
find that the performance of these approaches are not as good as \method.
Potential reasons for why these approaches do not work are (i) features that
differentiate benign and malicious traffic for \DO might not be features that
are important for \MT's model (as can be seen in the fact that for some \DOs,
\MT's model does not improve after data sharing) and (ii) the DecisionTree's
decision path when trained on \DO's data might be over-reliant on only one or
very few features, hence, do not provide useful signals to select $\Duseful$.

\begin{figure}[htbp]
  \centering
  \subfloat[CDF of \MT's F1 score after data sharing using \randsamp with $\Duseful$ budget of 5 samples for different classifiers. DecisionTree seems to be the best classifier for this dataset.]{
      \includegraphics[width=0.45\textwidth]{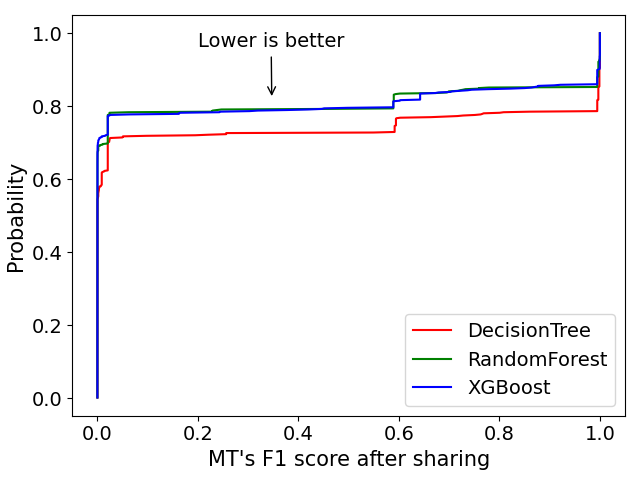}
 
  }
  \hfill
  \subfloat[CDF of \MT's F1 score after data sharing using \randsamp with $\Duseful$ budget of 100 samples for different classifiers. DecisionTree seems to be the best classifier for this dataset.]{
      \includegraphics[width=0.45\textwidth]{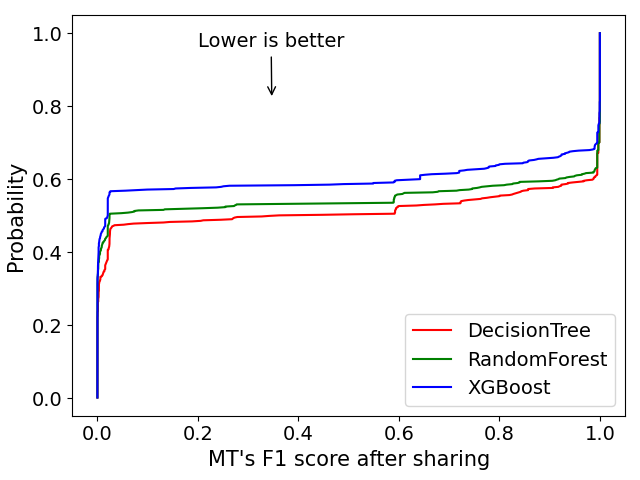}
     
  }
  \caption{Performance of \randsamp for different classifiers for the tabular dataset.}
  \label{fig:random_sampling_classifiers}
\end{figure}

\begin{figure}[htbp]
  \centering
  \subfloat[CDF of \MT's F1 score after data sharing using \method with $\Duseful$ budget of 5 samples for different classifiers. DecisionTree seems to be the best classifier for this dataset.]{
      \includegraphics[width=0.45\textwidth]{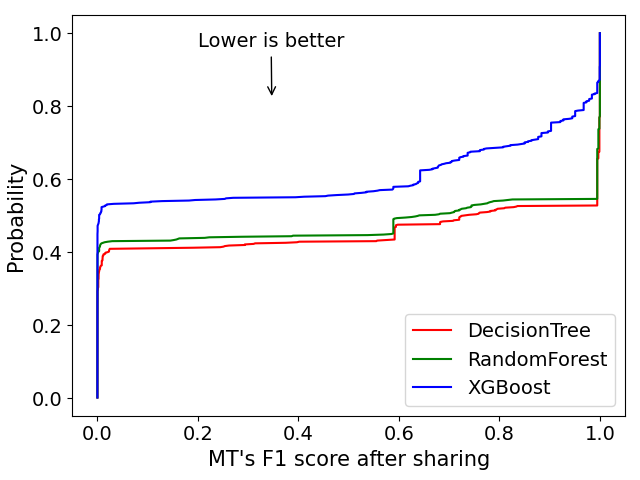}

  }
  \hfill
  \subfloat[CDF of \MT's F1 score after data sharing using \method with $\Duseful$ budget of 100 samples for different classifiers. DecisionTree seems to be the best classifier for this dataset.]{
      \includegraphics[width=0.45\textwidth]{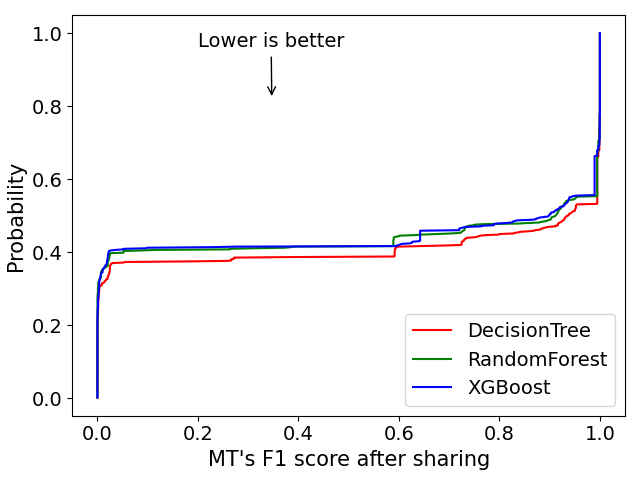}
     
  }
  \caption{Performance of \method for different classifiers for the tabular dataset.}
  \label{fig:Enola_classifiers}
\end{figure}

\begin{figure}[ht]
   \centering 
    \includegraphics[width=0.5\linewidth]{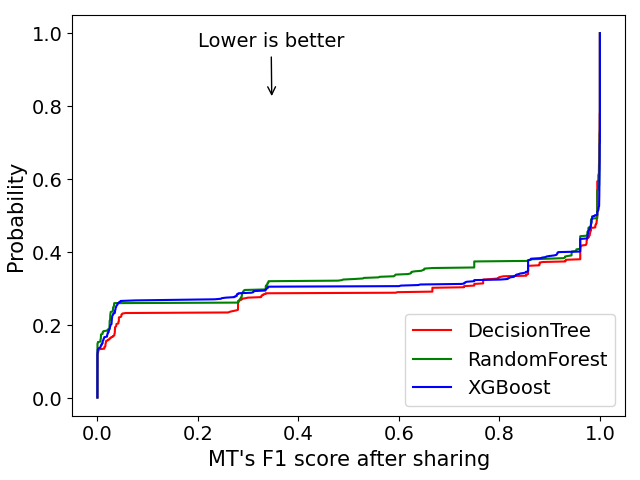}
    \caption{CDF of \MT's F1 score after data sharing for \fullinfo for different classifiers. DecisionTree seems to be the best classifier for this dataset.}\label{fig:full_info_classifiers}
  
\end{figure}

\textbf{Performance with different classifiers:} In this section, we present
\MT's F1 score after data sharing using \randsamp and \method for different
classifiers. We find that DecisionTree seems to be the best classifier for this
dataset. Refer to Figure \ref{fig:random_sampling_classifiers},
\ref{fig:Enola_classifiers} and \ref{fig:full_info_classifiers} for the results.

\begin{figure}[ht]
  \centering 
   \includegraphics[width=0.5\linewidth]{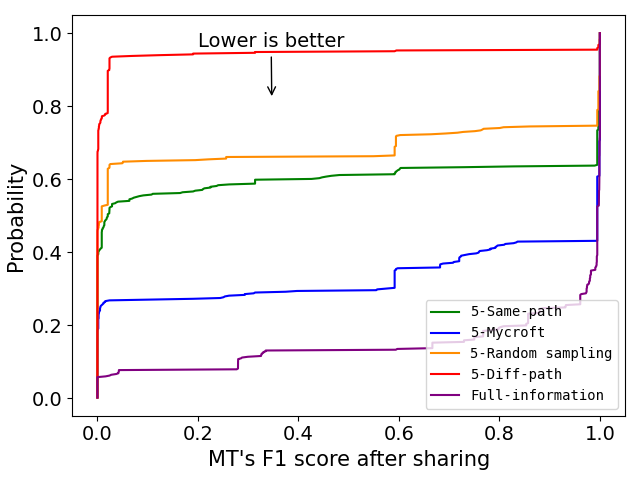}
   \caption{CDF of \MT's F1 score after data sharing using different data selection methods with $\Duseful$ budget of 5 samples and using \fullinfo. Classifier is DecisionTree. N = 466 cases where samples of the same decision path as $\Dhard$ and samples of different decision path as $\Dhard$ can be found.  }\label{fig:diff_sharing}
 
\end{figure}

\textbf{Performance when $\Duseful$ is selected based on different data
selection algorithm:} To explore whether DecisionTree's decision path can be
used to select $\Duseful$, we use $\Duseful$ budget of 5 samples and compare
\MT's F1 score after data sharing when $\Duseful$ is retrieved from samples of
the same decision path as $\Dhard$, of different decision paths as $\Dhard$,
retrieved from \randsamp and \method. Note that to make this study comparable,
we only consider cases where samples of the same decision path as $\Dhard$ and
samples of different decision path as $\Dhard$ can be found. This total up to
466 cases. We find that although sharing samples of same decision paths as
$\Dhard$ can be slightly better than \randsamp, \method still outperforms this
approach significantly. Refer to Figure \ref{fig:diff_sharing} for the
results.\\

\begin{figure}[ht]
  \centering 
   \includegraphics[width=0.5\linewidth]{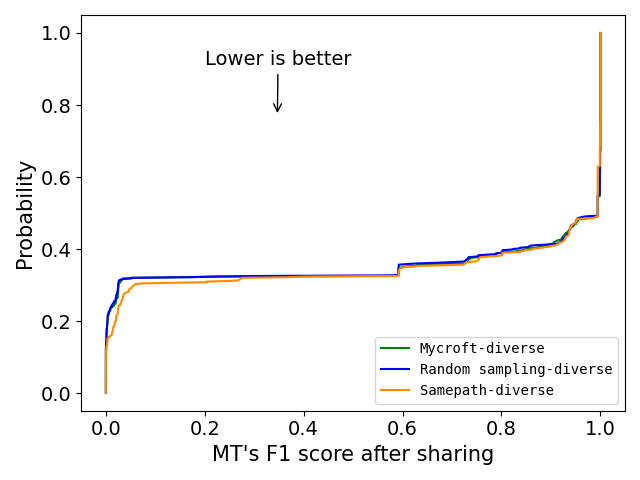}
   \caption{CDF of \MT's F1 score after data sharing when $\Duseful$ is made up of 100 samples retrieved from \method and 5 samples retrieved by different data selection methods. Classifier is DecisionTree. N = 574 cases where samples of the same decision path as $\Dhard$can be found. Note that the 5 samples later selected by each data selection method must be different from 100 samples already selected by \method.}\label{fig:diff_diversification}
 
\end{figure}

\textbf{Performance of combining \method and other data selection methods:}
To explore the effects of combining \method and other data selection methods
such as DecisionTree's decision path and \randsamp, we let $\Duseful$ to be made
up of samples selected by \method and samples selected by these other data
selection methods. We find that \MT's F1 score after data sharing is not
significantly improved when combining \method with other data selection methods
compared to using \method alone. Refer to Figure \ref{fig:diff_diversification}
for an example of the results.\\

\textbf{Performance using different distance metrics:} Depending on the type
of data (image versus tabular) and the dataset, different distance metrics might
be chosen to give better performance. For example, for the tabular dataset,
after experimenting with different distance metrics such as total variance
distance \cite{6804281}, autoencoder distance \cite{2020arXiv200608204A}, etc,
we found that existing distance metrics do not give good performance. As such,
we come up with a simple, intuitive distance metrics called ``binning distance''
to measure the distance between $\Dhard$ and \DO's samples. The intuition behind
this distance metrics is that because different features of tabular data has
different units and scales, we first bin the data (using histogram of uniform
distances) based on each feature's own distribution and then calculate the
distance between $\Dhard$ and \DO's samples based on the bins. Because the
distribution of the same feature might be different for \DO's data and $\Dhard$,
we need to based our binning method on the feature distribution of both sources
of data to ensure that the binning is not over-fitted to one source of data. As
such, we first create $D^{\text{base}}$ which is the union of $\Dhard$ and some
of \DO's samples and then bin the data based on the feature distribution of
$D^{\text{base}}$. We then calculate the distance between $\Dhard$ and \DO's
samples based on the bins. We find that this distance metrics gives better
performance compared to other distance metrics. Refer to Table
\ref{tab:binning_algo} for the algorithm to calculate the binning distance.

\section{Case studies} \label{appsec:case_studies}
Here, we present the details for the $\DO$s and $\MT$'s used for the Scenario 3 in Section~\ref{sec: case studies}. \\
\textbf{Image datasets:}\\
\texttt{DO-1}:  This $\DO$ contains the highest quantity of data from the same distribution as $\Dhard$ and is the same as the $\DO$ we use in other experiments involving 
the Dogs \& Wolves dataset. This $\DO$ should provide the highest utility to the $\MT$. \\
\texttt{DO-2}: \texttt{DO-2} is a noisy version of \texttt{DO-1} where we introduce noise by randomly transforming the images using PyTorch transforms~\cite{torchvisiontransforms}. 
The transforms we apply include Random crops, resizing, flipping, changing contrast and perspective. We expect the utility of training on such images to be lower as compared
to the clean images. \\
\texttt{DO-3}: \texttt{DO-3} has randomly sampled data from dog and wolf classes in the ImageNet dataset. We empirically verify that it contains a subset of data from the distribution
required by the $\MT$ and will thus be useful to the $\MT$ to some degree. \\
\texttt{DO-4}: This $\DO$ contains a small subset of the useful samples contained in \texttt{DO-1}. While useful in nature, this $\DO$'s ability to signal its utility should be limited. \\
\texttt{DO-5}: \texttt{DO-5} contains no data from the required training distribution and only consists of the data from $\MT$'s training distribution. This type of data should have
the least utility. \\

\textbf{Tabular datasets:}\\
In this case study, we focus on the \MT-7, which belongs to capture 20 and trying to predict the C\&C Torii attack. The goal of the \MT is to select, amongst 95 potential \DOs, the ones that will contain 
useful data that helps predicting this attack. If performing \fullinfo with \DOs, 88 out of 95 \DOs will give \MT an F1 score of > 0.99\% while the remaining 7 \DOs will give \MT an F1 score of <= 0.0\%.
Given that the utility of useful \DOs is almost the same, the goal of \MT-7 is is mostly to retrieve useful \DOs rather than rank them because any useful \DOs can improve \MT's performance.\\

We find that with a $\Duseful$ budget of only 5 samples, \method can retrieve 92 useful \DOs that can give \MT an F1 score of > 0.99\% (the rest gives \MT an F1 score of <= 0.0\%). This is much better compared to \randsamp which 
only retrieves 25 useful \DO that give \MT an F1 score of > 0.99\% (the rest give \MT an F1 score of < 0.0\%). Examining the cases where \method outperforms \fullinfo in selecting useful \DO, we find that \method can retrieve useful \DOs while \fullinfo is unable to because irrelevant data affects \fullinfo's ability to retrieve useful samples. For example,
there is a \DO created by mixing samples of PartOfAHorizontalPortScan attack from capture 3 with samples of C\&C attack from capture 3. For this \DO, \fullinfo will gives \MT an F1 score of 0.0\% while \method gives \MT an F1 score of 0.99\%. This is because although C\&C attack samples from capture 3 are useful for \MT (give \MT an F1 score of > 0.99\%) and PartOfAHorizontalPortScan attack samples from capture 3 are not (give \MT an F1 score of < 0.0\%).
\fullinfo will give \MT an F1 score of 0.0\% because the irrelevant PartOfAHorizontalPortScan attack samples from capture 3 will dominate useful C\&C attack samples data from capture 3. \method is able to identify this \DO as useful because it can select only the useful C\&C attack samples from capture 3 and ignore irrelevant PartOfAHorizontalPortScan attack samples from capture 3.